\documentclass{article} 
\usepackage{iclr2027_conference,times}
\usepackage{amsmath,amsfonts,bm}

\def\secref#1{section~\ref{#1}}

\def\eqref#1{equation~\ref{#1}}

\def\1{\bm{1}}

\DeclareMathAlphabet{\mathsfit}{\encodingdefault}{\sfdefault}{m}{sl}
\SetMathAlphabet{\mathsfit}{bold}{\encodingdefault}{\sfdefault}{bx}{n}

\usepackage{hyperref}
\usepackage{url}
\usepackage{xcolor}
\usepackage{array}
\usepackage{booktabs}
\usepackage{makecell}
\usepackage{multirow}
\usepackage{tabularx}
\usepackage{graphicx}
\usepackage{framed}

\usepackage{macros}

\title{The Hitchhikers Guide to Rubric Quality Understanding and Enrichment}

\author{
\textbf{Ankit Aich}$^{1}$, \enspace
\textbf{Zhengyang Qi}$^{1}$, \enspace
\textbf{Charles Dickens}$^{1}$, \enspace
\textbf{Derek Pham}$^{1}$, \enspace
\textbf{Esha Sharma}$^{1}$, \\
\textbf{Josh Viktorov}$^{1}$, \enspace
\textbf{Amanda Dsouza}$^{1}$, \enspace
\textbf{Armin Parchami}$^{1}$, \enspace
\textbf{Frederic Sala}$^{1,2}$, \enspace
\textbf{Paroma Varma}$^{1}$ \\
\vspace{3pt}
$^{1}$Snorkel AI \qquad
$^{2}$Department of Computer Sciences, University of Wisconsin--Madison \\
}
\iclrfinalcopy 
\begin{document}

\maketitle
\lhead{Preprint}

\begin{abstract}
    Rubrics distill notions of expert quality and measure agent performance. However, the quality of rubrics themselves have not been systematically measured and are often left to downstream performance.We import apparatuses from measurement theory built for exactly this: quantitative signals based on the rubric's content, and introduce the \textbf{RubrIc-Failure Taxonomy (RIFT)}, 
 of nine possible ways a rubric fails, organized under reliability and content validity. Every mode leaves a distinct signature. To show the signals track failure causally, we seed 720 corruptions, injecting each RIFT mode into clean rubrics at known severity levels. 
A linear probe over the signals identifies which mode was injected at $75.0\%$ accuracy, beating $56.7\%$ for a frontier model asked to name the failure directly. Surprisingly across GDPval and Terminal-Bench, 10 of 48 expert-authored rubrics weight their criteria backwards, putting more of the score on requirements an expert panel judged less essential. 
This means a response can fail what matters most and still be graded well.
This paper serves as a comprehensive guide on how to understand failure modes in rubrics and create better versions using quality signals, causal experiments, and provides a taxonomy with its rules and examples.

\end{abstract}

\commentout{Rubrics distill expert notions of quality into a single interpretable artifact that a model can apply on its own. The rubric itself, though, is not systematically checked. 
Quality is inferred largely from downstream evidence such as judge agreement or training outcomes, which conflates the rubric with the model applying it.
\emph{A rubric is a measurement instrument, but there is no established practice for measuring the quality of the rubrics themselves.} }

\section{Introduction}

Rubrics have become 
central to the
evaluation and training of large language models (LLMs) \citep{hashemi-etal-2024-llm, liu-etal-2026-openrubrics, ding2026adarubrictaskadaptiverubricsreliable}.
Prior work evaluates rubrics from downstream outcomes, including agreement rates between automated judges \citep{hashemi-etal-2024-llm, shen2026rethinkingrubricgenerationimproving} and improvements in models trained using rubric derived rewards \citep{gunjal2025rubrics, liu-etal-2026-openrubrics}. These methods conflate the rubric with the judge, and both are insensitive to the failure that matters most which is a rubric that is internally consistent, interpretable, but measuring the wrong thing. Recent work has begun to study rubric quality alongside downstream outcomes  \citep{arabzadeh-etal-2024-assessing, siro2026learningjudgellms, hong2026canllmswrite, roy2026premisepolicyrubricsmeasurement}.


Educational measurement and the social sciences have spent decades formalizing the validity of an assessment instrument. 
Most famously, Lawshe's content validity ratio for expert agreement on item essentiality \citep{lawshe1975quantitative}, and the decomposition of instrument validity into criterion, content, and construct validity \citep{straub2004validation, taherdoost2016validity}. 
Importing these into rubric evaluation yields quantifiable signals. 
However, signals alone tell half a story. 
Numeric signals can tell a practitioner that something is wrong; it does not tell them what is wrong, and therefore does not tell them what to fix. This necessitates an explanation that connects quality signals to specific failure modes.
\textbf{This paper supplies both, the quality signals and the taxonomy for explanations.}
\commentout{
What is missing is not another measurement, but a vocabulary that makes measurements \emph{sayable}.
\textbf{This paper supplies both halves and the mapping between them.}
}

\commentout{These are derived from the educational-measurement and assessment-design literature and organized under the three classical validity categories of \textbf{reliability, content validity, and consequential validity.} 
The taxonomy is not a detector. 
It is a naming layer i.e. a set of labels, each corresponding to a distinct and independently documented way that an assessment instrument can fail.}

To provide intuitive and measurable signals we propose four families of rubric quality signals. The first uses a panel of simulated experts to judge whether each criterion is essential to the task.
The second groups criteria by their relevance, from core requirements to peripheral concerns. The third evaluates how well criteria align with the task’s domain relevant objectives.
The fourth examines whether criterion weights reflect their importance.
Together, these four families provide computable signals for assessing rubric quality.

We also introduce \textbf{RIFT}, the \textbf{R}ubr\textbf{I}c \textbf{F}ailure mode \textbf{T}axonomy, a vocabulary to label rubric quality issues with \textbf{nine} rubric failure modes: \textbf{subjective language, ungrounded verification, non-atomicity, missing criteria, misalignment, rigidity, redundancy, and hackability, and miscalibrated weighting}. 
We map quality signals to the taxonomy empirically and causally with known corruptions: \textbf{720} injected rubric variants across \textbf{60} expert-human-authored, weighted rubrics drawn from two agentic benchmark families GDPVal \citep{patwardhan2025gdpval} and Terminal-Bench \citep{merrill2026terminalbench}, spanning three corruption levels. The injected mode is known by construction in every case so we can ask directly whether the psychometric signature of a rubric identifies the way it is broken.
Every mode produces a statistically distinguishable signature: \textbf{46 of 136} mode-by-signal cells are significant under BH-FDR correction, with strong effect sizes up to $|d|=2.83$. Together these help practitioners identify, quantify, and label rubric failures early in the development process, avoiding expensive failures stemming from leaving problem discovery solely in downstream tasks. 

\begin{figure}[t]
    \centering
    \setlength{\fboxsep}{0pt}
    \fbox{%
        \includegraphics[
            width=\dimexpr\linewidth-1\fboxrule\relax
        ]{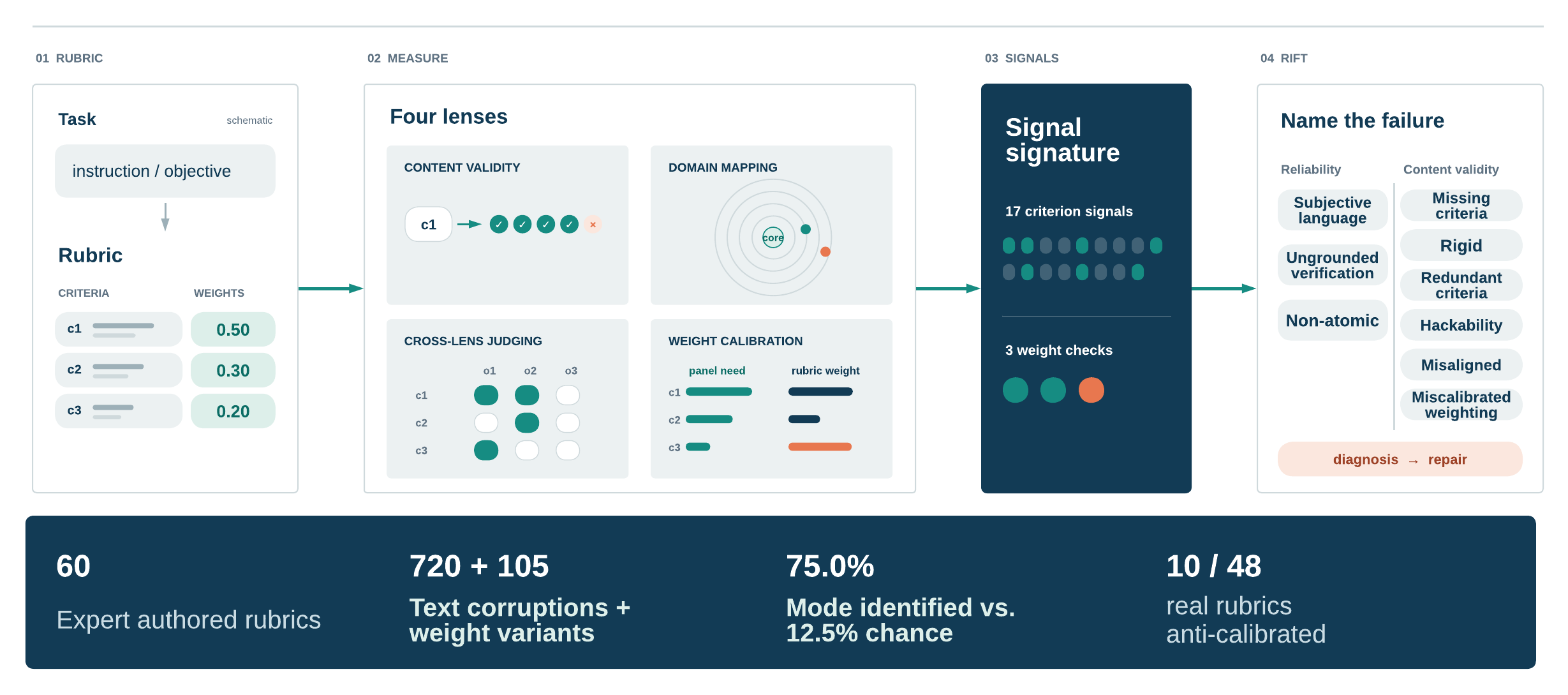}%
    }
    \caption{This paper grounds rubric design in well-established quality signals. These help us understand and view causal effects on rubric design. We also provide a taxonomy \textbf{(RIFT)} to help designers name specific failure modes.}    \label{fig:placeholder}

\end{figure}


\commentout{\paragraph{The main contributions of this paper are: }
\begin{itemize}

    \item \textbf{A formal set of quantifiable quality signals:} to understand rubric failures. Sec \S\ref{sec:quality_signals} 
    
    \item \textbf{A failure-mode taxonomy for rubrics:} derived from measurement theory. Sec \S\ref{sec:rift} 
    
    \item \textbf{Causal experiments turning the numeric signal to a label-able failure mode :} Sec \S\ref{sec:results} 
    
    \commentout{\item \textbf{An empirical mapping from signals to named failures:} Section \S\ref{sec:results}}

\end{itemize}

This paper provides its readers with knowledge curated from education and measurement literature for the better design of rubrics. 
It also provides readers a vocabulary of derived failure modes and a mapping to quantifiable signals so future designers of rubrics can practically measure the quality of rubrics while also being able to name the deficiencies grounded in scientific rigor. }

\textbf{This paper makes three contributions}. We define a set of quantifiable quality signals computed directly from a rubric's content (\S\ref{sec:quality_signals}), derive a taxonomy of rubric failure modes from measurement theory (\S\ref{sec:rift}), and establish through causal experiments that the signals identify which failure mode is present rather than only that something is wrong (\S\ref{sec:results}). 
Together these give a rubric designer two things they did not have: a way to measure a rubric before it is used, and a taxonomy, grounded in decades of assessment research, for naming what the measurements say is wrong.
\section{Related Work}
\label{sec:background}

\subsection{Rubrics in the era of agents}

\commentout{Research on rubrics in the era of LLMs falls broadly into three areas: constructing and refining rubrics, applying them to evaluate LLM outputs, and, most recently, assessing the quality of the rubrics themselves.}

\commentout{\paragraph{Constructing and refining rubrics.} A first line of work builds rubrics and the models that generate them. \citet{liu-etal-2026-openrubrics} curate prompt-rubric pairs for training rubric-generation and rubric-based reward models, and \citet{research_rubrics_scale} construct a rubric-centered dataset targeting competencies such as instruction following and research ability. \citet{shen2026rethinkingrubricgenerationimproving} decompose coarse criteria into finer, more discriminative ones, while \citet{ding2026adarubrictaskadaptiverubricsreliable} generate rubrics adaptively from task descriptions to score agent trajectories and produce reward signals for preference learning. \citet{dhole2026rubricraginterpretablereliablellm} ask whether such rubrics are interpretable to human users, and find that off-the-shelf LLM rubrics diverge substantially from human-authored ones.

\paragraph{Applying rubrics to evaluation.} A second line uses rubrics as the evaluation mechanism itself. \citet{hashemi-etal-2024-llm} assess natural language text along dimensions specified by a manually constructed rubric, and \citet{wei2025conceptbasedrubricsimprovellm} show that concept-based rubrics narrow the gap between LLM judges and supervised models on STEM student responses, and further seed synthetic training data for lightweight evaluators.

\commentout{\citet{qi2026riftrubricfailuremodetaxonomy} develop a grounded taxonomy of rubric failure modes spanning reliability, content-validity, and consequential-validity failures, together with automated diagnostics that detect these failure modes directly from a rubric's structure rather than from downstream judge behavior.}

\paragraph{Assessing rubric quality.} A third and more recent line asks whether a rubric is itself \emph{good}. Rubric quality has historically been inferred indirectly, from judge-human agreement or downstream training outcomes, both of which conflate the rubric with the judge applying it. \citet{arabzadeh-etal-2024-assessing} take an early step toward isolating it, verifying that criteria are essential, robust, and discriminative through stability checks and adversarial deletion. Subsequent work measures rubric quality along several axes: rubric-level meta-evaluation \citep{pan2026rubricevalrubriclevelmetaevaluation}, scoring reliability and cross-model consistency \citep{siro2026learningjudgellms}, and intrinsic and extrinsic agreement with human-authored references \citep{hong2026canllmswrite}. \citet{roy2026premisepolicyrubricsmeasurement} treat rubrics as measurement specifications and audit them along four axes, cautioning that agreement between judges is not validity, and that fitting human preferences is not robustness to gaming. Together this work establishes that rubrics can be deficient. It does not supply a vocabulary for saying how.}

Rubrics are used to evaluate and guide the training of AI agents.
Recent research investigates rubric development methodologies \citep{liu-etal-2026-openrubrics, research_rubrics_scale, shen2026rethinkingrubricgenerationimproving, ding2026adarubrictaskadaptiverubricsreliable}, applies them as the evaluation mechanism itself \citep{hashemi-etal-2024-llm, wei2025conceptbasedrubricsimprovellm}, and, most recently, asks whether a rubric is itself \emph{good} \citep{arabzadeh-etal-2024-assessing, pan2026rubricevalrubriclevelmetaevaluation, siro2026learningjudgellms, hong2026canllmswrite, roy2026premisepolicyrubricsmeasurement, dhole2026rubricraginterpretablereliablellm}. 
For instance, that third line of research finds machine-generated rubrics diverging from human-authored ones and warns that agreement between judges is not validity \citep{dhole2026rubricraginterpretablereliablellm}. 
Together this line of work demonstrates that rubrics can yield reliable downstream measurements but lack validity. 
\commentout{
Together it establishes that rubrics can be deficient; it does not supply a vocabulary for saying how. 
Recent LLM work has no principled account of the ways rubrics fail.
Evaluation has also been focused on downstream effects. 
}
Rubrics come from educational measurement, a field has spent decades formalizing measurement, labels, and computable statistics. In this paper we bridge these fields, providing practitioners a systematic account of the ways rubrics fail how to holistically quantify their quality before observing downstream effects. 

\commentout{\citet{qi2026riftrubricfailuremodetaxonomy} develop a grounded taxonomy of rubric failure modes spanning reliability, content-validity, and consequential-validity failures, together with automated diagnostics that detect these failure modes directly from a rubric's structure rather than from downstream judge behavior.}

\subsection{The history and design of rubrics: lessons outside CS}
\label{subsec: rubrics_outside_cs}

Rubrics were studied as measurement instruments long before they were used to evaluate LLMs. Educational assessment distinguishes analytic rubrics, which score criteria one at a time and support formative feedback, from holistic rubrics, which collapse everything into a single judgment that is faster to apply but less informative \citep{brookhart2018appropriate}, and examines what makes descriptive rubrics effective in practice \citep{brookhart2015quality}.

A rubric is a measurement instrument. So the broader validity literature applies to it directly. Validity concerns how well an instrument covers the domain under investigation \citep{ghauri2005research} and whether it measures what it is intended to measure \citep{field2005discovering}. \citet{taherdoost2016validity} decomposes it into criterion, face, content, and construct validity. Of these, content validity, the degree to which items reflect the content universe the instrument generalizes to \citep{straub2004validation}, is the most directly computable: \citet{lawshe1975quantitative} estimates it numerically from panel agreement on whether each item is essential, and consensus methods extend the approach to full questionnaires \citep{montes2023design}.


\section{Quality Signals}
\label{sec:quality_signals}

We define four families of quality signals, each importing an established validation measure and answering a different question. \emph{Content validity} asks whether each criterion belongs in the rubric. \emph{Domain mapping} asks whether the criterion set covers the domain in the right proportions. \emph{Cross-lens judging} asks whether criteria align with the task's objectives. \emph{Weight calibration} asks whether scoring mass sits on the criteria that matter. 
Together, these four families provide a holistic view of rubric quality.

A rubric is a collection of $M$ \emph{criteria} $\{(c_i, w_i)\}_{i=1}^{M}$, each pairing a natural-language statement $c_i$ with a non-zero numerical \emph{weight} $w_i$ expressing its importance. 
Weights may be negative where a rubric penalizes undesired behavior. 
We write $s_i = |w_i| / \sum_k |w_k|$ for the \emph{scoring mass} a criterion commands, so that $\sum_i s_i = 1$.

\subsection{Content validity ratio}
\label{sec:exp-cvr}

CVR is Lawshe's \emph{content validity ratio} (CVR) \citep{lawshe1975quantitative}, which is still active recently \citep{almanasreh2019evaluation}.
A panel of $N$ experts labels each criterion essential or not, and the content validity ratio for criterion $i$ is
\begin{equation}
    \mathrm{CVR}_i \;=\; \frac{n_i^{e} - N/2}{N/2} \;\in\; [-1, 1],
    \label{eq:cvr}
\end{equation}
where $n_i^{e}$ is the number of essential votes for criterion $i$.
The statistic reaches $+1$ under unanimous endorsement, sits at $0$ at a bare majority, and turns negative when most of the panel judges the criterion inessential. We report a rubric's mean CVR, following the adjacent content-validity-index tradition's caution to state the aggregation explicitly \citep{polit2006content}.

A panel is instantiated with $N{=}5$ task-adapted personas, generated by a GPT-5.5 model per task from the task instruction alone. 
Each persona is a one-line expert profile covering a distinct specialism relevant to that task. 
An example for a prehospital cardiac-care task, the generated set comprises an EMS medical director, a cardiac QA paramedic, an interventional cardiologist, a protocol compliance officer, and a clinical documentation auditor. 
Examples can be found in Appendix \ref{app:personas}

\subsection{Domain mapping}
\label{sec:exp-domain}

Content validity evaluates criteria one at a time. Domain mapping evaluates the set as a distribution. Instrument-development protocols require an explicit content plan before criteria are judged \citep{fives2013tos, kalkbrenner2021measure}, and validity theory supplies the two questions such a plan must answer: content \emph{relevance} \S\ref{sec:exp-cvr}, and content \emph{representativeness}, whether the set covers the domain in the right proportions \citep{messick1989validity, haynes1995content, sireci1998content}. Alignment methodology turns these into gradeable criteria with explicit thresholds \citep{webb1997criteria, webb2007alignment, porter2002content}, while a fourth tradition places items in a domain \emph{geometry} via expert ratings and multidimensional scaling \citep{sireci1992analyzing, sireci1995smes}, where radial distance from the center is itself measurement-bearing structure \citep{guttman1954radex, marshalek1983radex}. To our knowledge no existing rubric-validation framework combines these elements; our domain mapping automates all four.

For each task a GPT-5.5 model derives from the instruction a \emph{domain map}: five ordered relevance levels, \textsc{core}, \textsc{adjacent}, \textsc{background}, \textsc{general}, and \textsc{irrelevant}, each populated with task-specific terms. 
Let $d_i \in \mathcal{D}$ denote the relevance level assigned to
criterion $c_i$, where
$\mathcal{D} = \{\text{\textsc{core}}, \text{\textsc{adjacent}}, \text{\textsc{background}},
\text{\textsc{general}}, \text{\textsc{irrelevant}}\}$.
The rubric's domain alignment is the fraction of
criteria assigned to each level:
\begin{equation}
    p_r = \frac{1}{M}\sum_{i=1}^{M}\mathbf{1}[d_i = r],
    \qquad r \in \mathcal{D},
    \label{eq:domain-mapping}
\end{equation}
where $\mathbf{1}[\cdot]$ is the indicator function.
For e.g. in the prehospital cardiac audit, \textsc{core} covers ECG timing compliance while \textsc{general} covers report structure and prose. 
We record per-level threshold indicators and flag any criterion landing at \textsc{irrelevant}. 

\commentout{Expert rubrics concentrate sharply, with mean $p_{\textsc{core}}{=}0.73$ and $p_{\textsc{adjacent}}{=}0.15$, so mass drifting outward marks a rubric measuring something other than its task.}

\subsection{Cross-lens expert judging}
\label{sec:exp-crosslens}

Following structured elicitation \citep{dalkey1963experimental}, $E{=}5$ expert personas independently propose objectives from the instruction. 
These consolidate into a list of objectives carrying normalized weights $W_j$ and one of four centrality tiers 
\citep{raymond2001job, lawshe1975quantitative}, with an LLM judge \citep{zheng2023judging} scoring panel agreement and objective verifiability. Each expert rates every criterion against every objective on a $0$--$3$ scale, the multidimensional form of item--objective congruence \citep{rovinelli1976content, turner2003indexes}, giving $\bar{R}_{ij}\in[0,1]$.

Classical construct threats \citep{messick1989validity} then become computable. 
\emph{Item relevance} $\frac{1}{M}\sum_i\max_j\bar{R}_{ij}$ and \emph{weighted relevance} $\frac{1}{M}\sum_i\sum_j W_j\bar{R}_{ij}$ catch criteria tied to no objective, or only to peripheral ones. \emph{Representativeness} $1-\mathrm{JSD}(W\|A)$, with realized coverage $A_j\propto\sum_i s_i\bar{R}_{ij}$ and base-2 Jensen--Shannon divergence \citep{lin1991divergence}, measures construct underrepresentation; \emph{contamination}, the $s_i$-weighted mass on background and irrelevant tiers, construct-irrelevant variance \citep{haladyna2004construct}; \emph{redundancy}, the mean cosine among criterion pairs at similarity $\geq0.8$ (zero if none), duplicated content \citep{boyle1991item, yen1993scaling}. Higher is better for the first three, worse for the last two.

\textbf{Discriminant controls.} We build a nomological net \citep{cronbach1955construct} and close it discriminantly \citep{campbell1959convergent}. Three items from each of four classes are scored identically to the rubric's own criteria: expert positives, adjacent-but-wrong-domain near negatives, generic quality statements, and far negatives. We report the margins separating candidate criteria from the generic and near-negative classes. A valid rubric must separate its criteria from plausible impostors, not merely score them highly, a check aimed at known LLM-judge biases \citep{zheng2023judging, salewski2023incontext}.

\subsection{Weight calibration}
\label{sec:exp-weight}

\commentout{The three families above all read criterion text. None examines the weight vector $w$, and} A rubric whose every criterion is sound can still misrepresent the construct if $w$ allocates scoring mass to the wrong criteria. Because practice-analysis methods derive numeric blueprint weights directly from expert judgment \citep{raymond2001job, lawshe1975quantitative}, a rubric's weights are themselves a testable claim about relative importance rather than a scoring convenience. \commentout{Three statistics test that claim against the same panel already used for $\mathrm{CVR}_i$ (Eq.~\ref{eq:cvr}). so no additional annotation is required.} \emph{Ordering calibration} asks whether $|w_i|$ ranks criteria the way the panel does, via the Spearman rank correlation
\begin{align}
    \rho_{\mathrm{ord}} = \mathrm{Spearman}\big(\{|w_i|\}_{i=1}^M,
    \{\mathrm{CVR}_i\}_{i=1}^M\big) \in [-1, 1],
    \label{eq:ordcvi}
\end{align}

undefined by convention when either vector is constant. Values near $1$ indicate weights that track panel necessity, while negative values indicate \emph{anti-calibration}: the rubric awards more credit for satisfying the less essential criterion of a pair.

Let $\mathcal{C} = \{i : \mathrm{CVR}_i > 0\}$ denote the panel-endorsed \emph{core} criteria. \emph{Failure score retention} (FSR) is the worst-case score fraction a response retains after failing exactly one core criterion while satisfying every other:
\begin{align}
    \mathrm{FSR} = \max_{i \in \mathcal{C}} \left(1 - \frac{|w_i|}
    {\textstyle\sum_{k: w_k > 0} w_k - \sum_{k: w_k < 0} w_k}\right),
    \label{eq:fsr}
\end{align}

which reduces to $1 - \min_{i \in \mathcal{C}} s_i$ under all-positive weights. High FSR flags rubrics where failing a core criterion is cheaply compensated by credit elsewhere. Relatedly, \emph{core weight share} $\sum_{i \in \mathcal{C}} s_i$ reports what fraction of total scoring mass the panel-endorsed criteria command at all. Higher ordering calibration and core weight share are better; higher FSR is worse.
\section{RIFT: The RubrIc Failure Mode Taxonomy}
\label{sec:rift}

\begin{figure}
    \centering   \includegraphics[width=0.9\linewidth]{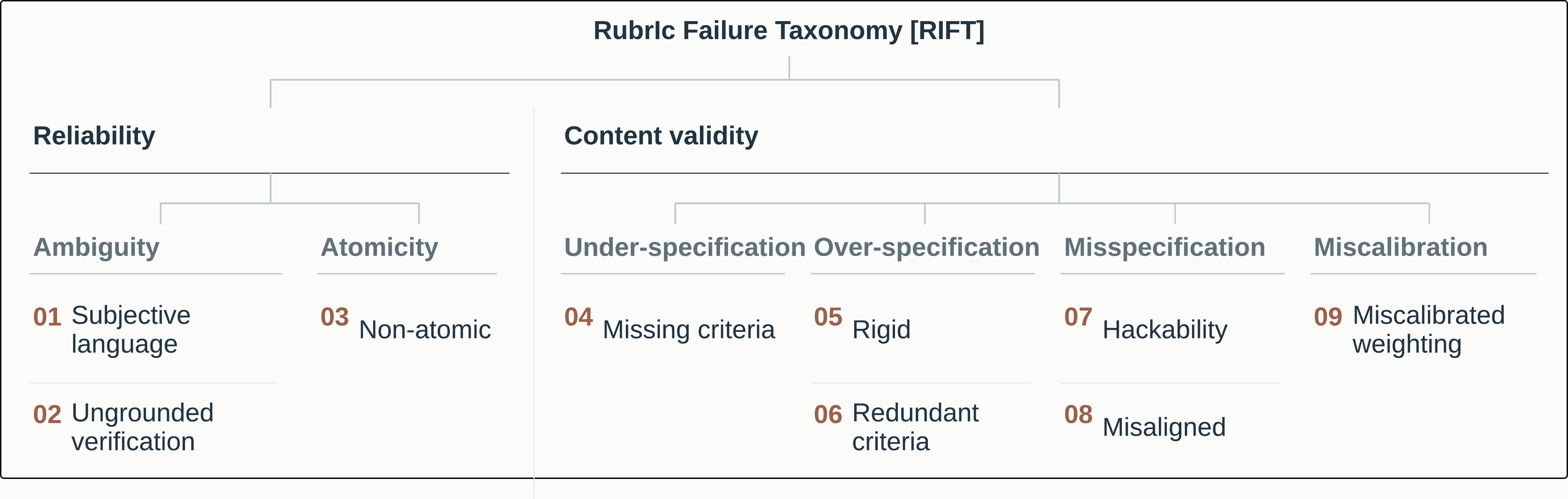}
    \caption{The RubrIc Failure Taxonomy (RIFT) organizes nine modes of rubric failure.
    See table \ref{tab:rift_taxonomy} for complete definitions.}  \label{fig:rift_taxonomy_fig}
\end{figure}

We introduce the \textbf{RubrIc Failure mode Taxonomy (RIFT)}, a general framework for identifying and classifying common rubric failures, and the vocabulary we use to explain \emph{how} those failures move the quality signals of \secref{sec:quality_signals}. RIFT is hierarchical: quality \emph{dimensions} organize \emph{sub-dimensions}, which group individual \emph{failure modes}. Each mode carries a complete definition, decision rules specifying when it applies, and pass and fail examples that fix its boundaries. Figure \ref{fig:rift_taxonomy_fig} previews the structure and \appref{app:rift} gives the full taxonomy.

\commentout{
}

\subsection{Grounding RIFT in assessment research}

A rubric is a measurement instrument, and measurement theory distinguishes two properties of one. \textit{Reliability} is the reproducibility of a rubric's evaluations. \textit{Content Validity} is the extent to which its content supports the intended interpretation of those evaluations. Reliable evaluations are necessary but not sufficient for valid interpretations~\citep{messick1989validity,cook:ajm06}.

Research on descriptive rubrics emphasizes clear, focused, substantive, and purpose-appropriate criteria~\citep{brookhart2015quality,brookhart2018appropriate}, which RIFT captures through the Reliability sub-dimensions of \textit{Ambiguity} and \textit{Atomicity}. Content-validity research defines quality by whether an instrument's items are relevant to and representative of its intended domain~\citep{lawshe1975quantitative,haynes1995content,sireci1998content}, distinguishing omissions that under-represent the intended content from extraneous requirements and proxies that introduce construct-irrelevant variance~\citep{messick1989validity,haladyna2004construct}, and identifying rephrased near-duplicates as a source of redundancy~\citep{boyle1991item}. RIFT extends these into the Content Validity sub-dimensions of \textit{Under-Specification}, \textit{Over-Specification}, \textit{Misspecification}, and \textit{Miscalibration}. \appref{app:rift_development} details the corpus, annotation rounds, LLM-assisted critique aggregation, and expert refinement. 

\commentout{Within this structure, the individual failure modes and their decision rules were derived from expert critiques through grounded analysis and comparative refinement~\citep{glaser:book67}. RIFT thus pairs an organizing structure grounded in established measurement concerns with failure modes grounded in observed defects in AI evaluation rubrics. \appref{app:rift_development} details the corpus, annotation rounds, LLM-assisted critique aggregation, and expert refinement.}

\subsection{Intervention operators}
\label{sec:causal_validation_of_rift}

To support our causal analysis in \secref{sec:results}, each failure mode is realized by an \emph{operator} that applies it to a clean rubric at a requested corruption rate $\rho$, so the clean rubric serves as its own control and the treatment effect is the difference between the original and the corrupted rubric. 
Criterion-level operators sample a $\rho$ fraction of criteria to delete, duplicate, merge, or rewrite.
The Miscalibrated Weighting operator is deterministic, moving a $\rho$ fraction of scoring mass from panel-endorsed core criteria to non-core criteria.
Full operator definitions and prompts are provided in \appref{app:corruption_prompts}.

\section{Experiments and Results}
\label{sec:results}

 We establish a causal connection between RIFT and quality signals with four main RQs:
\begin{itemize}
    \item RQ1) \textit{does each mode leave a distinct signature }(\S\ref{sec:res-signatures})?,
    \item RQ2) \textit{do quality signals degrade with corruption level }(\S\ref{sec:res-dose})?,
    \item RQ3) \textit{do the findings transfer across rubric families and the signals carry independent information} (\S\ref{sec:res-mediation})? and
    \item RQ4) \textit{what do criterion-reading signals not see }(\S\ref{sec:res-weight})?
\end{itemize}

All four questions need rubrics with known defects. 
We start from clean expert-authored rubrics and use the intervention operators of \secref{sec:causal_validation_of_rift} to inject RIFT failure modes at controlled severity. 
We score each clean and corrupted rubric on the quality signals so any movement in the signals is attributable to the mode we injected.

\textbf{Corpus and injection.} 
We use $T{=}60$ human-authored rubrics, 30 from
GDPval \citep{patwardhan2025gdpval} and 30 from
Terminal-Bench \citep{merrill2026terminalbench}.
We apply each of RIFT's eight criterion-level operators at $\rho{=}0.5$ across all 60 tasks (480 variants) and at $\rho\in\{0.25,0.75\}$ on a seeded 15-task subset (240 variants) for corruption-level analysis. 
The ninth mode, Miscalibrated Weighting, uses the deterministic weight-level operator and is evaluated separately in \S\ref{sec:res-weight}. Feature scoring uses GPT-5.5 for the baseline rubric and every variant alike, while corruption is authored by a sonnet-4-6 model, so scorer and corruptor are independent. 
Task-level instruments (personas, domain map, objective blueprint, discriminant controls) are fixed per task across arms, so within-task deltas isolate the injected mode. 
Appendix~\ref{app:cross_judge} replicates the main results under additional scoring models to show that our results are not model-specific artifacts.

\subsection{Injected failure modes produce mode-specific signatures}
\label{sec:res-signatures}

Each mode  damages a rubric in its own recognizable way. Across the $8 \text{ mode} \times17 \text{ quality signal}$ grid, \textbf{46 of 136 cells} are FDR-significant ($q<.05$), and every mode moves at least one signal at $\rho{=}0.5$ ($n{=}60$/mode; Figure~\ref{fig:heatmap}, per-mode effect sizes in Appendix~\ref{app:signatures}).

\begin{figure}[t]
    \centering
    \setlength{\fboxsep}{0pt}
    \fbox{%
        \includegraphics[
            width=\dimexpr\linewidth-1\fboxrule\relax
        ]{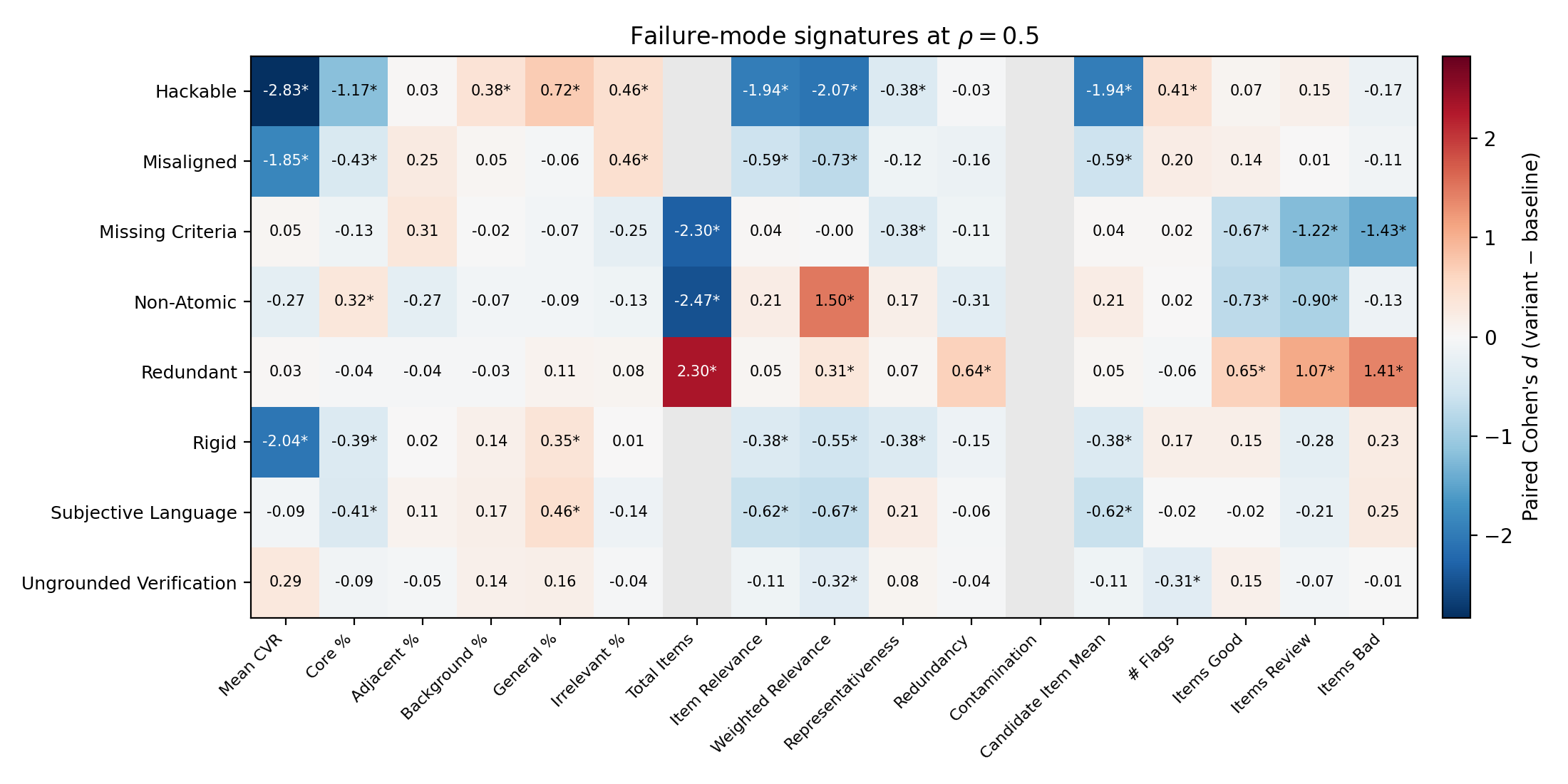}%
    }
    \caption{Failure-mode signatures at $\rho{=}0.5$: paired Cohen's $d$ (variant $-$
    baseline) for each of 8 injected modes across 17 signals, $n{=}60$ tasks per mode.
    Asterisks mark cells significant under Benjamini--Hochberg FDR at $q<.05$ applied
    across the full $8\times17$ grid (46 of 136). Blank cells are signals whose delta
    was identically zero on every task.}
    \label{fig:heatmap}
\end{figure}

The misspecification modes degrade content validity. 
\emph{Hackable} most severely of all at $d{=}{-2.83}$ on mean CVR ($\text{AUC}{=}0.13$). 
The modes affecting structure (redundant, non-atomic, missing criteria) move counts by $4$ to $7$ criteria while content validity holds. One entry runs against type: non-atomic \emph{raises} weighted relevance ($d{=}{+}1.50$), because merging criteria makes the survivors look more relevant.

\textbf{The signals are selective, not one severity dial.}
Content validity fires for the three misspecification modes while item-count signals fire for the three structural modes. Redundancy score fires for redundant alone, which is a tight signal-to-construct correspondence. Only weighted relevance behaves as a general severity dial, moving for 7 of 8 modes.

\textbf{The signature identifies the mode and the damage.} 
A multinomial logistic classifier over the 17-signal delta vector recovers which mode was injected with \textbf{75.0\% accuracy} (macro-F1 $0.746$\commentout{; chance $12.5\%$; $n{=}480$}), under leave-one-task-out cross-validation with all 8 arms of a held-out task removed together. 
Only two confusion pairs are large: subjective-language with ungrounded-verification under Ambiguity, and misaligned with rigid. 
Collapsing just these pairs raises accuracy to $89.4\%$. 
The classifier was never told the taxonomy has sub-dimensions, yet it recovered them from the signals alone. Adding Miscalibrated Weighting as a ninth label costs little (\textbf{72.0\%}, macro-F1 $0.712$, chance $11.1\%$), with mode-9 recall $0.68$. Our validation compares each corrupted rubric against its own clean parent. The signals themselves are absolute, as the anti-calibration finding demonstrates, so diagnosis outside this setting requires population-level norms in place of a per-rubric reference.

\textbf{A judge cannot do this by inspection}. We showed the same 540 units (480 corrupted, 60 clean) to an LLM judge given the instruction and the rubric, and asked it to name one of the eight modes or ``none.'' The judge, \texttt{claude-opus-4-8}, scores \textbf{56.7\%} (macro-F1 $0.530$), \textbf{18.3 points} below the feature-based classifier. \commentout{and it comes from a different provider than the GPT-5.5 scorer behind our features, so the gap is not an artifact of shared model idiosyncrasies.} Its failures are specific: it over-predicts misaligned (recall $1.00$, precision $0.33$), recognizes healthy rubrics only $18\%$ of the time, and gets zero of 60 ungrounded-verification units right.

\subsection{Signals grade severity, not just presence}
\label{sec:res-dose}

Knowing a rubric is damaged is less useful than knowing how badly. On the 15-task subset we regress each signal on the realized corruption fraction per unit. The modes that damage content validity grade cleanly: mean CVR falls at $-1.55$ per unit realized $\rho$ for hackable ($p{=}1.6\times10^{-12}$), $-1.35$ for rigid, and $-1.10$ for misaligned, each replicating independently in both datasets. Non-atomicity grades just as reliably in the opposite direction ($+0.159$, $r{=}{+}.67$), confirming that relevance inflation scales with merging rather than being an artifact of one corruption level. This matters beyond our setting: any quality measure built on mean per-criterion relevance can be improved by consolidating criteria.\commentout{Two modes do not grade at all, redundancy saturating at every level (slope $-0.005$, n.s.) and ungrounded verification staying near-invisible throughout ($-0.018$, n.s.).} Content validity therefore acts as a \emph{dosimeter} for misspecification while the domain and cross-lens signals largely act as \emph{presence detectors}, and a practitioner should read a depressed CVR as an estimate of severity but a fired detector only as a flag for inspection. See (Appendix~\ref{app:dose}) for detailed visual results.

\subsection{Effects transfer across rubric families, and the signals are independent}
\label{sec:res-mediation}

Rubric family moderates six of the eight criterion-level modes but never reverses their sign. Deliverable-style GDPval rubrics degrade more sharply than Terminal-Bench's technical checklists under every content-validity mode ($\Delta$mean\_CVR $-0.87$ vs.\ $-0.42$ for rigid, $-0.80$ vs.\ $-0.61$ for hackable, $-0.65$ vs.\ $-0.45$ for misaligned; all $q<.02$, BH-corrected within the eight moderation tests), plausibly because prose criteria admit more room for the corruption to work. The one reversal is mechanical: Terminal-Bench rubrics are longer, so deleting a fixed fraction removes more criteria. \commentout{Non-atomic and redundant show no family difference at all, with near-identical effects in both ($p \ge .57$). Thresholds should therefore be calibrated per family while the qualitative signature holds unchanged.}

If the signals were three views of one latent quality factor, the damage one registers would be explained by another. A bootstrap mediation analysis on hackable (2000 resamples) finds the CVR collapse routed neither through the domain shift ($0\%$ mediated via $\Delta$core\_pct; CI $[-.094, +.066]$) nor through relevance loss ($2\%$ via $\Delta$item\_relevance, CI straddling zero), against a total effect of $-0.700$. The signals register the same damage through independent channels, showing diagnostic power lives in the full vector and not in one scalar. 

\subsection{The blind spot: weight miscalibration}
\label{sec:res-weight}
So far we have dealt with criterion \emph{text}, and every criterion-level operator preserves total rubric weight, so a rubric can pass all of them while allocating scoring mass to the wrong criteria. We close this gap using weight-calibration statistics of \S\ref{sec:exp-weight}.

\textbf{Anti-calibration is present in human-authored rubrics.} Scored against the real $N{=}5$ panel on all 60 rubrics, expert weights modestly track panel-rated necessity on average (mean $\rho_{\mathrm{ord}}{=}{+}0.195$, $t{=}4.44$, $p{=}5.5\times10^{-5}$, $n{=}48$ scoreable). There still remains a substantial minority: \textbf{10 of 48 scoreable rubrics are negative}, ranking criteria against panel-rated necessity, so the rubric awards more credit for the less essential criterion of a pair. This is the Miscalibrated Weighting phenotype in expert-authored rubrics rather than injected. The remaining 12 rubrics have a constant CVR vector, admitting no correlate-able ranking, so we do not score them to preserve the statistic's meaning.

\begin{figure}[t]
\centering
\includegraphics[width=\textwidth]{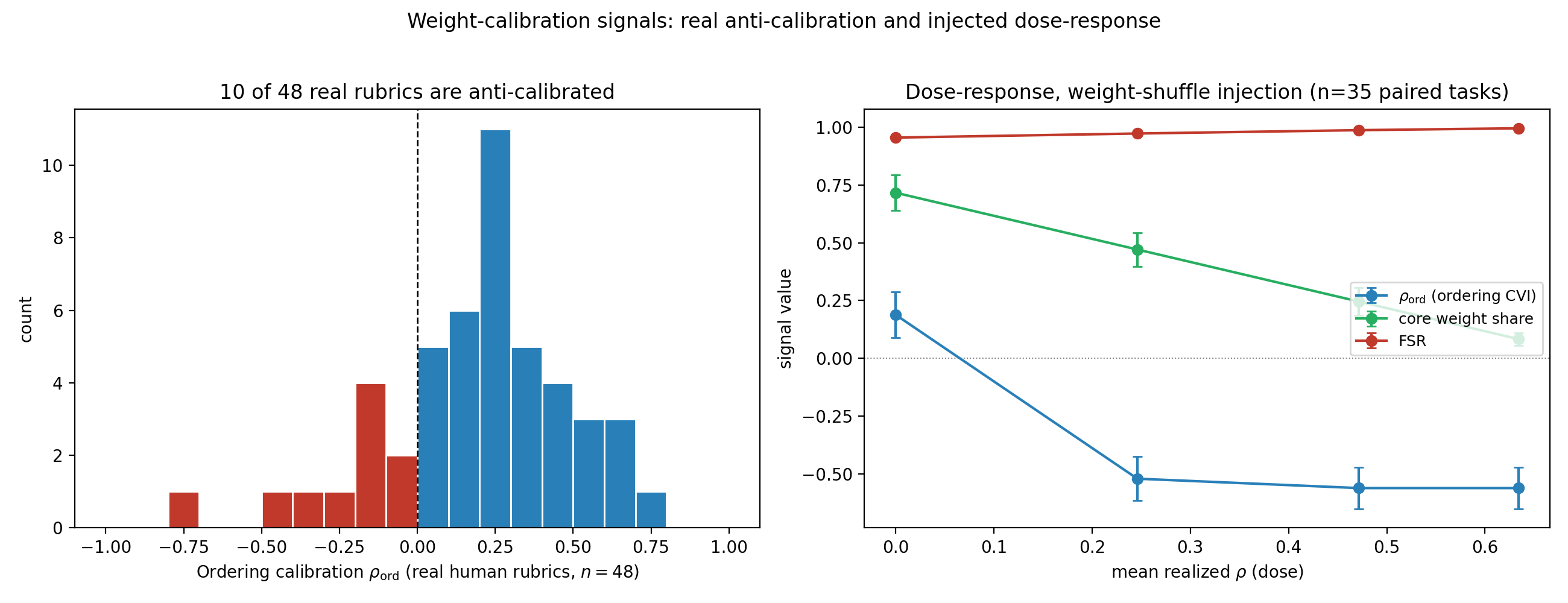}
\caption{Left: distribution of $\rho_{\mathrm{ord}}$ over the 60 human-authored rubrics ($n{=}48$ scoreable); red bars are anti-calibrated. Right: the three weight-calibration signals under weight-shuffle injection, mean $\pm95\%$ CI over 35 paired tasks against mean realized $\rho$.}
\label{fig:weightcal}
\end{figure}

\begin{table}[t]
\centering
\small
\caption{Weight-calibration signals on human rubrics ($n{=}60$, panel-scored) and under weight-shuffle injection ($n{=}35$ paired). Higher $\rho_{\mathrm{ord}}$ and core weight share are better; higher FSR is worse.}
\label{tab:weightcal}
\begin{tabular}{@{}lrrrrr@{}}
\toprule
& \textbf{Human} & \textbf{Clean} & $\rho{=}.25$ & $\rho{=}.50$ & $\rho{=}.75$ \\
\midrule
$\rho_{\mathrm{ord}}$ & $+0.195$ & $+0.189$ & $-0.522$ & $-0.562$ & $-0.562$ \\
Core weight share & $0.835$ & $0.717$ & $0.471$ & $0.246$ & $0.083$ \\
FSR & $0.950$ & $0.956$ & $0.974$ & $0.988$ & $0.996$ \\
\bottomrule
\end{tabular}
\end{table}

\textbf{Injection confirms all three statistics respond} (Table~\ref{tab:weightcal}). On 105 corrupted rubrics across 35 paired tasks, $\rho_{\mathrm{ord}}$ flips sign at the first corruption level ($+0.19\to-0.52$, $d{=}{-1.98}$, $p{=}1.8\times10^{-13}$), core weight share falls monotonically ($0.72\to0.08$, all $p{<}10^{-21}$), and FSR rises monotonically ($0.956\to0.996$, all $p{<}10^{-8}$). This is correct since moving mass off core criteria makes a core failure cheaper to absorb. The three differ in \emph{how} they grade. Within the corrupted arms alone, core weight share and FSR remain graded ($r{=}{-}0.43$ and $r{=}0.45$, both $p<10^{-5}$), tracking how much mass has moved, while $\rho_{\mathrm{ord}}$ saturates ($r{=}0.005$) and reports only whether the move changed the ranking. The operator's positivity constraint caps movement at high $\rho$, so realized rather than requested $\rho$ is the correct axis throughout.

\textbf{The criterion-reading signals fail exactly as predicted.} Running the same 105 units through the 17-signal grid, 16 of 17 stay silent and the one that fires, \texttt{representativeness\_score}, is the only signal whose formula consumes weights. Its response is continuous and dose-graded (slope $-0.078$, $p{=}6.2\times10^{-4}$, $n{=}105$), monotone across bins. This validates our mechanism. The grid responds exactly where the measurement theory says weights should enter, which is evidence the signals measure what they claim. It also shows why the dedicated statistics are needed: representativeness can tell you the weighting is wrong, but not which criteria are at fault.
\section{Discussion}
\label{sec: discuss}

In this section we discuss each major takeaway and its interpretation. 

\textbf{From measurement to diagnosis.} The central claim of this work is  that psychometric signals identify \emph{how} a rubric is damaged. A linear probe over the 17-signal vector recovers the injected failure mode at $75.0\%$ accuracy against a $12.5\%$ chance baseline, and does so under leave-one-task-out validation, so the mapping is not task-specific. It exceeds a strong LLM asked directly to name the failure mode by $18.3$ points ($56.7\%$), indicating that the measurement apparatus contributes information that judge intuition does not already supply. \commentout{And its errors are structured rather than diffuse: the two dominant confusion pairs are precisely the taxonomy's sibling dimensions, and collapsing them raises accuracy to $89.4\%$. A vocabulary derived from measurement theory and an instrument derived from measurement theory, developed against different constraints, converge on the same partition of rubric failure. That convergence is the strongest evidence we can offer that the taxonomy names real distinctions rather than convenient ones.}

\textbf{The instrument measures what it claims to measure.} Two results speak to construct validity from opposite directions. The redundancy signal fires for injected redundancy and for none of the other modes, giving a signal-to-construct correspondence that is rare in evaluation metrics. More strikingly, the miscalibrated-weighting mode leaves every item's text byte-identical and perturbs only the weight vector; $16$ of the $17$ signals remain silent, and the one signal that fires is the only one whose computation consumes weights at all. Its response is dose-graded on the realized corruption fraction (slope $-0.078$, $p=6.2\times10^{-4}$, $n=105$), with a monotone gradient across dose bins. \commentout{The instrument's sensitivity is located exactly where its measurement theory predicts it should be. Read together with the mediation analysis, in which the collapse of content validity under the hackable mode is routed neither through domain shift nor through relevance loss (proportion mediated $0.00$ and $0.02$ respectively), the picture is of three lenses registering damage through independent channels rather than three views of one latent quality factor. Diagnostic power resides in the vector, not in any summary score.}

\textbf{A division of labor across families.} The quality signals are not interchangeable, and their differences are functional. CVR behaves as a dosimeter for the misspecification modes, falling at $-1.55$ per unit corruption fraction for hackability ($p=1.6\times10^{-12}$) \commentout{with the effect replicating independently within both benchmark families.} The domain-mapping lens behaves as a presence detector \commentout{: it registers that misspecification has occurred but grades its severity only weakly (slope $-0.23$, $r=-0.35$).} The cross-lens panel is what catches the modes that invert the logic of the other two. \commentout{The redundancy detector illustrates a third behavior, saturating rather than grading, firing at every dose without distinguishing among them.} A practitioner reading these signals should therefore read them for different purposes: some answer whether, some answer how much, and some answer only which.

\textbf{Relevance metrics can be inflated by consolidation.} The non-atomicity result deserves emphasis because it generalizes beyond our setting. Merging several criteria into one compound item makes the rubric worse by every substantive standard, yet mean weighted relevance \emph{rises} ($d=+1.50$), and rises further as more items are merged (dose slope $+0.159$, $r=+0.67$). The mechanism is not subtle: a consolidated item touches more of the task, so per-item relevance scores it higher. Any rubric-quality metric built on averaged per-item relevance is therefore optimizable by consolidation, and any rubric-generation procedure that treats such a metric as a reward will drift toward compound, unatomic criteria. This is a concrete instance of the general hazard the taxonomy is meant to surface, and it recommends that mean-relevance measures always be reported alongside item counts.

\textbf{Implications of RIFT.} Readers are now able to name deficiencies in grounded theory. Previously, confronting a low content validity ratio and a depressed core-relevance share pointed to a faulty rubric. The mapping established here licenses a stronger statement: that the rubric is probably hackable, and that the indicated repair is to bind its items to task-specific evidence. Rubric family moderates degradation severity, with deliverable-style rubrics degrading more sharply than technical checklists across six of eight headline comparisons, though direction never reverses. So thresholds should be calibrated per family while the qualitative signature transfers unchanged.

\section{Conclusion} 

Rubric quality has been gauged through downstream model performance conflating the rubric with the model applying it. We give practitioners two things instead. First, a set of quantified signals that read a rubric directly, drawn from decades of assessment and measurement research. The second is a taxonomy for naming what those signals find. We validated both by seeding known defects into expert-authored rubrics and recovering which defect was injected from the signals alone. Our results are causal and we show in appendix \ref{app:cross_judge} that these are not the artifact of a single model or family. 

Two limitations follow from our methods. We isolated one failure mode at a time, while rubrics in practice fail in subtler and overlapping ways. We measured absolute weight magnitude, not the penalties, gates, and non-additive scoring rules that some rubrics also use.

The experiment this most immediately motivates is to derive reward signals from clean and corrupted rubrics and compare what the resulting policies learn. While our research points to the MECE (mutual exclusive and collectively exhaustive) nature of our taxonomy, we also encourage readers to always upgrade the evaluations to which we hold other evaluation instruments.

\section*{AI disclosure section}

In this work, we used generative AI tools for coding assistance and help with grammatical editing and formatting of the manuscript. We have not used generative AI tools for literature review, writing of the content of the paper, formulation of the research problems and question, or selection of the quality signals. 
The use of AI in the creation of the RIFT taxonomy is disclosed in \appref{app:rift}. 
The rest of the required disclosure tasks are not applicable to this work.  We have reviewed all AI-assisted work. Experiments, results, code, citations, and structure, were verified and tested for correctness by the authors. We take responsibility for the final content of this work, including text, claims or artifacts produced with the aid of generative AI.

\bibliography{iclr2027_conference}
\bibliographystyle{iclr2027_conference}

\appendix
\section{Appendix}
\subsection{RIFT Taxonomy Development Process}
\label{app:rift_development}

RIFT was developed using a grounded-theory methodology and constant-comparative refinement~\citep{glaser:book67}. Rather than beginning from a fixed rubric-quality schema, we iteratively derived failure modes from expert critiques of real rubrics. 
The development corpus contained 85 rubrics and 255 expert annotations drawn from five sources spanning human-authored and synthetic rubric-generation settings. 
The human-authored sources were \textsc{AdvancedIF}~\citep{he2025advancedifrubricbasedbenchmarkingreinforcement}, which provides expert-authored criteria for complex instruction following, and \textsc{ResearchRubrics}~\citep{research_rubrics_scale}, which focuses on factual grounding and multi-step analytical reasoning for deep research agents. 
The synthetic sources were \textsc{WildChecklists}~\citep{viswanathan2025checklistsbetterrewardmodels}, \textsc{OpenRubrics}~\citep{liu-etal-2026-openrubrics}, and \textsc{AutoRubrics}~\citep{anonymous2026autorubric}, which cover checklist-style requirements, contrastive rubric generation, and reward-oriented rubric induction.

The 85 rubrics were divided into four subsets and processed in four sequential refinement rounds. 
In the first round, three experts independently wrote open-ended \textit{rubric critiques} for each rubric without access to a predefined taxonomy. 
These critiques were randomly shuffled, batched in groups of five, and provided to GPT-5.2 to propose an initial set of candidate failure modes. 
The LLM was used only as an aggregation and proposal mechanism: all proposed modes, names, definitions, and boundaries were manually reviewed and revised by the expert panel before being used in subsequent rounds.

In later rounds, experts labeled the next subset of rubrics using the current taxonomy and assigned all applicable failure-mode labels. 
They also wrote additional rubric critiques for failures not covered by the current taxonomy and \textit{taxonomy critiques} for problems in the taxonomy itself, such as unclear definitions, overlapping labels, missing distinctions, or structural inconsistencies. 
After each round, the panel reviewed the collected labels and critiques, refined the taxonomy, and then applied the updated version to the next subset. 
The process terminated when the panel agreed that additional rubrics were no longer producing new failure modes or substantive taxonomy revisions.

The resulting taxonomy, summarized in \tabref{tab:rift_taxonomy}, organizes rubric failures into two dimensions. 
\textit{Reliability} captures failures that make rubric judgments inconsistent or difficult to reproduce across graders. 
\textit{Content Validity} captures failures in the relationship between rubric criteria and the intended evaluation target, including missing, overly narrow, misaligned, gameable, redundant, or misweighted criteria.

\section{RIFT: RubrIc Failure Mode Taxonomy}
\label{app:rift}

\begin{table}[t]
    \centering
    \footnotesize
    \setlength{\tabcolsep}{4pt}
    \renewcommand{\arraystretch}{1.1}
    \caption{RIFT: RubrIc Failure Mode Taxonomy. Nine failure modes under two quality dimensions. Full definitions, decision rules, and pass/fail examples in Appendix~\ref{app:rift}.}
    \label{tab:rift_taxonomy}

    \begin{tabularx}{\textwidth}{@{}llX@{}}
        \toprule
        \textbf{Dimension} & \textbf{Sub-dimension} & \textbf{Failure mode} \\
        \midrule

        \multirow{3}{*}{Reliability}
        & \multirow{2}{*}{Ambiguity}
          & \textbf{Subjective Language}: evaluative terms left unanchored to objective expectations \\
        & & \textbf{Ungrounded Verification}: checkable facts with no reference, source, or procedure \\
        \cmidrule(l){2-3}
        & Atomicity
          & \textbf{Non-Atomic}: bundled requirements prevent consistent partial credit \\
        \midrule

        \multirow{6}{*}{Content Validity}
        & Under-Specification
          & \textbf{Missing Criteria}: a prompt requirement that no criterion evaluates \\
        \cmidrule(l){2-3}
        & \multirow{2}{*}{Over-Specification}
          & \textbf{Rigid}: stricter or narrower requirements than the prompt asks for \\
        & & \textbf{Redundant Criteria}: one behavior scored by two or more criteria \\
        \cmidrule(l){2-3}
        & \multirow{2}{*}{Misspecification}
          & \textbf{Hackability}: top score reachable by inflating proxies \\
        & & \textbf{Misaligned}: grades the wrong objective or embeds incorrect assumptions \\
        \cmidrule(l){2-3}
        & Miscalibration
          & \textbf{Miscalibrated Weighting}: secondary mass offsets failing the core requirement \\
        \bottomrule
    \end{tabularx}
\end{table}

\commentout{\begin{table*}[t]
    \centering
    \small
    \setlength{\tabcolsep}{5pt}
    \renewcommand{\arraystretch}{1.15}
    \caption{RIFT: RubrIc Failure Mode Taxonomy. Full definitions, decision rules, and pass/fail examples in Appendix~\ref{app:rift}.}
    \label{tab:rift_taxonomy}

    \begin{tabularx}{\textwidth}{@{}
        >{\centering\arraybackslash}m{0.11\textwidth}
        >{\centering\arraybackslash}m{0.13\textwidth}
        >{\centering\arraybackslash}m{0.14\textwidth}
        >{\raggedright\arraybackslash}X
        @{}}
        \toprule
        \textbf{Dimension} & \textbf{Sub-dimension} & \textbf{Failure mode} & \textbf{Definition preview} \\
        \midrule

        \multirow{3}{=}{\centering Reliability}
        & \multirow{2}{=}{\centering Ambiguity}
        & \makecell[c]{Subjective\\Language}
        & Uses subjective evaluative terms (``clear,'' ``appropriate'') without anchoring them to objective expectations. \\

        &
        & \makecell[c]{Ungrounded\\Verification}
        & Requires checking facts that are in principle verifiable, but supplies no reference, source, time scope, or procedure for doing so. \\
        \cmidrule(lr){2-4}

        & Atomicity
        & Non-Atomic
        & Bundles multiple requirements into one criterion, preventing consistent partial credit. \\
        \midrule

        \multirow{6}{=}{\centering Content\\Validity}
        & \makecell[c]{Under-\\Specification}
        & \makecell[c]{Missing\\Criteria}
        & The prompt implies a checkable requirement that no criterion evaluates. \\
        \cmidrule(lr){2-4}

        & \multirow{2}{=}{\centering Over-Specification}
        & Rigid
        & Imposes stricter or narrower requirements than the prompt asks for. \\

        &
        & \makecell[c]{Redundant\\Criteria}
        & Two or more criteria evaluate the same requirement, scoring one behavior repeatedly. \\
        \cmidrule(lr){2-4}

        & \multirow{2}{=}{\centering Misspecification}
        & Hackability
        & Gameable: a top score is reachable by inflating proxies without improving correctness. \\

        &
        & Misaligned
        & Grades the wrong objective for the prompt, or embeds incorrect assumptions. \\
        \cmidrule(lr){2-4}

        & Miscalibration
        & \makecell[c]{Miscalibrated\\Weighting}
        & Secondary criteria carry enough mass to offset failing the core requirement. \\
        \bottomrule
    \end{tabularx}
\end{table*}
}

This section provides the complete descriptions of each failure mode in the RIFT taxonomy (\tabref{tab:rift_taxonomy}), including detailed decision rules for annotation.
Each failure mode includes: (1)~when to apply the label, (2)~how to determine whether a rubric exhibits the failure mode, and (3)~boundary conditions specifying when \emph{not} to apply the label and which alternative label to consider instead.
For each failure mode, we also provide illustrative pass examples (rubrics that do \emph{not} exhibit the failure mode) and fail examples (rubrics that \emph{do} exhibit the failure mode).
Note: in the ``Do NOT apply'' cross-references below, CSV-internal label identifiers have been replaced with the display names used in \tabref{tab:rift_taxonomy}.

\subsection{Reliability Failures}

Reliability failures lead to inconsistent grading across annotators or evaluation runs, reducing the reproducibility and trustworthiness of rubric-based evaluation.

\paragraph{Subjective.}
\label{fm:subjective}
Apply when the rubric uses inherently subjective evaluative terms (e.g., ``clear,'' ``appropriate,'' ``credible,'' ``comprehensive,'' ``professional,'' ``engaging,'' ``well-written,'' ``good sources'') and does NOT sufficiently anchor them with objective expectations.

\textbf{How to determine:}
\begin{itemize}
    \item Identify criteria dominated by inherently subjective terms.
    \item Check whether the rubric provides ANY anchoring attempt such as:
    \begin{itemize}
        \item concrete checklists (``includes X/Y/Z''),
        \item measurable thresholds (word count, required sections, required elements),
        \item examples/anti-examples of what qualifies vs does not qualify, or
        \item explicit decision rules (``count as clear if it defines the term and gives one example'').
    \end{itemize}
    \item If the rubric relies primarily on grader judgment and provides no meaningful anchors, apply.
\end{itemize}

\textbf{Important clarification:}
\begin{itemize}
    \item Do NOT apply if the rubric gives examples or non-trivial decision rules that explain what the subjective term means (even if the term is still somewhat subjective).
\end{itemize}

\textbf{Do NOT apply if:}
\begin{itemize}
    \item The core issue is missing expected answers/tolerances or a bounded verification procedure for a groundable requirement (use \emph{Ungrounded}).
    \item The requirement is entirely absent (use \emph{Missing Criteria}).
\end{itemize}

\textbf{Illustrative examples:}

\begin{small}
\begin{tabularx}{\linewidth}{@{} l X X @{}}
\toprule
& \textbf{Input Context} & \textbf{Rubric} \\
\midrule
\textsc{Pass} & Write a professional email declining a meeting. & \textit{2 pts: Includes a decline + proposes an alternative time. 2 pts: Uses a greeting and sign-off. 1 pt: No negative or insulting language.} \\
\addlinespace
\textsc{Fail} & Summarize the study. & \textit{10 pts: The summary is clear and sufficiently detailed.} \\
\bottomrule
\end{tabularx}
\end{small}

\paragraph{Non-Atomic.}
\label{fm:non_atomic}
Apply when the rubric does not provide a parseable, consistently scorable structure OR uses bundled (non-atomic) criteria that prevent consistent partial credit.

\textbf{Triggers} (any sufficient):

\textit{Non-atomic (bundled) criteria}
\begin{itemize}
    \item One scored item bundles multiple independently scorable requirements with no partial-credit rule or separable sub-scores (e.g., ``clear, comprehensive, accurate, and well-cited'' as a single 10-pt item).
\end{itemize}

\textbf{Do NOT apply when:}
\begin{itemize}
    \item Subparts are separately scored or the rubric provides explicit level anchors (e.g., ``1 point each for A/B/C'' or a 0--2 scale per dimension with definitions).
    \item The rubric is scorable but uses subjective language (use \emph{Subjective}) or is missing requirements (use \emph{Missing Criteria}).
\end{itemize}

\textbf{Illustrative examples:}

\begin{small}
\begin{tabularx}{\linewidth}{@{} l X X @{}}
\toprule
& \textbf{Input Context} & \textbf{Rubric} \\
\midrule
\textsc{Pass} & Write a short answer with two supporting reasons. & \textit{2 pts: Answers the question. 1 pt: Reason \#1 supports the answer. 1 pt: Reason \#2 supports the answer. 1 pt: Total length <=150 words.} \\
\addlinespace
\textsc{Fail} & Summarize the article. & \textit{10 pts: Summary is clear, accurate, comprehensive, concise, and engaging.} \\
\bottomrule
\end{tabularx}
\end{small}

\paragraph{Ungrounded.}
\label{fm:ungrounded}
Apply when the rubric requires verification that is plausibly groundable/boundable, but the rubric does not provide the necessary grounding (answer keys/acceptable variants/tolerances/decision rules) OR does not bound the verification procedure (what to check, how much to check, and how to judge conflicts).

\textbf{How to determine} (any sufficient):
\begin{enumerate}
    \item[(A)] \textit{Groundable determinate tasks lack grading anchors.} The task has a knowable target output given fixed inputs (e.g., extraction, classification, translation, math, SQL result, code output), but the rubric provides no expected answers, acceptable variants, label mappings, tolerances, or decision rules. The criterion may be clearly worded (e.g., ``totals are correct''), yet graders still lack what they need to check correctness.
    \item[(B)] \textit{Open-world requirements lack bounded audit procedure.} The rubric demands broad verification (e.g., ``all facts are true,'' ``restaurants are open right now,'' ``fully original/no plagiarism,'' ``links work'') without bounding: what to check (scope/sample size), which sources/tools are allowed, how to resolve conflicting evidence, and the pass/fail threshold.
    \item[(C)] \textit{Measurement standard is unspecified but could be made checkable.} The rubric requires a measurement that depends on an unspecified standard (e.g., ``exactly 20 pages'') without defining the rendering/formatting standard or offering a workable proxy.
\end{enumerate}

\textbf{Do NOT apply if:}
\begin{itemize}
    \item The requirement is simply missing from the rubric (use \emph{Missing Criteria}).
    \item The main issue is subjective wording without anchors (use \emph{Subjective}). The rubric provides a representative list of examples to demonstrate expected content.
\end{itemize}

\textbf{Illustrative examples:}

\begin{small}
\begin{tabularx}{\linewidth}{@{} l X X @{}}
\toprule
& \textbf{Input Context} & \textbf{Rubric} \\
\midrule
\textsc{Pass} & Extract all email addresses from the text. & \textit{1 pt per correct email address; accepted forms include plus-addressing. Gold list of emails: a@x.com, b.y@z.org, \ldots\ Deduct 1 pt per missing email.} \\
\addlinespace
\textsc{Fail} & Compute the correct totals for these 30 invoices. & \textit{10 pts: Totals are correct.} \\
\bottomrule
\end{tabularx}
\end{small}

\subsection{Content Validity Failures}

Content validity failures arise when rubric criteria are misaligned with the intended evaluation target, either by grading the wrong objective or by failing to cover essential requirements.

\paragraph{Misaligned or Rigid.}
\label{fm:misaligned}
Apply when the rubric (a) grades the wrong objective for the prompt or embeds incorrect assumptions, OR (b) imposes unnecessarily strict/narrow requirements not asked for by the prompt or reasonably inferred from the prompt, predictably penalizing prompt-faithful high-quality answers.

\textbf{How to determine} (any applies):
\begin{itemize}
    \item Wrong task / shifted objective: makes non-requested deliverables mandatory for points.
    \item Incorrect embedded assumptions: assumes a context not in the prompt (jurisdiction, audience, tools, constraints) and scores accordingly.
    \item Penalizes good practice: scores down reasonable caveats/uncertainty/safety practices when the prompt does not forbid them.
    \item Arbitrary brittleness/over-constraint: mandates a specific tool/library/method/structure/formatting or false precision when multiple reasonable alternatives would satisfy the prompt.
\end{itemize}

\textbf{Do NOT apply when:}
\begin{itemize}
    \item The prompt itself imposes the strictness at any point (e.g., exact JSON keys, or a direct instruction from the user earlier in a chat conversation).
    \item The requirement is missing entirely (use \emph{Missing Criteria}).
    \item The main problem is internal contradiction (use \emph{Self-Contradictory}).
    \item The main problem is rubric-level proxy gaming (use \emph{Hackable}).
\end{itemize}

\textbf{Illustrative examples:}

\begin{small}
\begin{tabularx}{\linewidth}{@{} l X X @{}}
\toprule
& \textbf{Input Context} & \textbf{Rubric} \\
\midrule
\textsc{Pass} & Write Python code to parse CSV. & \textit{5 pts: Correct parsing. 3 pts: Handles quoted commas. 2 pts: Includes brief usage example. (Does not mandate pandas vs csv module.)} \\
\addlinespace
\textsc{Fail} & Write a haiku about winter. & \textit{5 pts: Includes at least 5 academic citations. 5 pts: Uses APA format reference list.} \\
\bottomrule
\end{tabularx}
\end{small}

\paragraph{Missing Criteria.}
\label{fm:missing}
Apply when the prompt implies at least one checkable must-have requirement, but the rubric provides no criterion that allows a grader to evaluate that requirement at all.

\textbf{How to determine:}
\begin{enumerate}
    \item List the prompt's core requirements:
    \begin{itemize}
        \item required deliverables/components,
        \item must/must-not constraints,
        \item required format/ordering/sections,
        \item and genre-critical qualities the prompt clearly expects (e.g., functional correctness for code; ``two sentences''; ``valid JSON''; ``include 10 items''; ``chronological order'').
    \end{itemize}
    \item For each requirement, check whether ANY rubric criterion covers it.
    \item If one or more requirements have no corresponding criterion, apply.
\end{enumerate}

\textbf{Do NOT apply if:}
\begin{itemize}
    \item The rubric mentions the requirement but is vague or subjective (use \emph{Subjective}).
    \item The rubric mentions the requirement but it cannot be graded consistently due to missing keys/tolerances/bounded audit steps (use \emph{Ungrounded}).
    \item The rubric grades a different task or adds arbitrary constraints (use \emph{Misaligned or Rigid}).
\end{itemize}

\textbf{Illustrative examples:}

\begin{small}
\begin{tabularx}{\linewidth}{@{} l X X @{}}
\toprule
& \textbf{Input Context} & \textbf{Rubric} \\
\midrule
\textsc{Pass} & Return ONLY valid JSON with keys: name (string) and age (integer). & \textit{3 pts: Output parses as JSON. 2 pts: Contains exactly keys name and age. 2 pts: name is a string; age is an integer. 1 pt: No surrounding commentary.} \\
\addlinespace
\textsc{Fail} & Write a 200-word email and include a subject line. & \textit{10 pts: Tone is professional. 5 pts: Grammar and spelling are correct.} \\
\bottomrule
\end{tabularx}
\end{small}

\paragraph{Miscalibrated Weighting.}
\label{fm:miscalibrated}
Apply when the rubric's point distribution does not reflect the relative importance of requirements in the prompt. A response can earn an implausibly high score despite failing a core task requirement, or lose disproportionate credit for a secondary requirement.

\textbf{How to determine:}
\begin{enumerate}
    \item \textbf{Core-failure retention:} Identify a requirement essential to completing the prompt, then calculate how much score a response can retain if it completely fails that requirement while satisfying everything else.
    \item \textbf{Compensating credit:} Check whether points from secondary, formatting, or supporting criteria can offset failure on the primary task objective.
    \item \textbf{Relative importance:} Compare each criterion's weight to its importance in the prompt. Core requirements should materially affect the achievable score; secondary requirements should not dominate it.
    \item \textbf{Whole-rubric sanity check:} Compare a substantively successful answer with one that misses a major requirement. The rubric should produce a meaningful score difference between them.
\end{enumerate}

\textbf{Do NOT apply if:}
\begin{itemize}
    \item The criterion evaluates the wrong objective or assumes context not present in the prompt (use \emph{Misaligned}).
    \item The objective is appropriate, but the criterion demands one unnecessarily narrow method or answer path (use \emph{Rigid}).
    \item A necessary requirement is not evaluated anywhere in the rubric (use \emph{Missing Criteria}).
    \item The problem is that one criterion combines multiple independently scorable requirements (use \emph{Non-Atomic}).
    \item The rubric can be gamed through proxy signals or superficial compliance, regardless of its numerical weights (use \emph{Hackability}).
    \item A low-weight requirement is genuinely secondary to the prompt, even if an external reference or gold answer treats it as important.
\end{itemize}

\textbf{Illustrative examples:}

\begin{small}
\begin{tabularx}{\linewidth}{@{} l X X @{}}
\toprule
& \textbf{Input Context} & \textbf{Rubric} \\
\midrule
\textsc{Pass} & Write a decision memo on whether to freeze Palisade for external review. & \textit{5 pts: Recommends holding the freeze because the merge-time guardrail failed. 3 pts: Reports the 128 KiB median error. 1 pt: Correct filename.} \\
\addlinespace
\textsc{Fail} & Write a decision memo on whether to freeze Palisade for external review. & \textit{5 pts: Correct filename. 5 pts: Reports the 128 KiB median error. 5 pts: Clear headings. 1 pt: Recommends holding the freeze.} \\
\bottomrule
\end{tabularx}
\end{small}

\subsection{Consequential Validity Failures}

Consequential validity failures reduce the downstream usefulness and discriminative power of rubric-based evaluation, even when individual criteria may be well-defined.

\paragraph{Hackable.}
\label{fm:hackable}
Apply when the rubric is gameable at the rubric level: a responder could easily achieve a top score by inflating proxy metrics (length, number of bullets/sections/items/citations/examples/brands, repeated keywords) without materially improving correctness, relevance, or fulfillment of the prompt---and the rubric lacks strong quality gates that tie points to substantive, prompt-aligned success.

\textbf{Core question} (required):
\begin{itemize}
    \item Could I easily achieve full marks on this rubric while still not satisfying the prompt requirements or producing a low-quality response?
\end{itemize}

\textbf{How to determine} (any sufficient):
\begin{itemize}
    \item Most points come from ``more'' ($\geq$N tips/citations/examples/pros/cons/sections) while relevance, non-duplication, correctness, and prompt-specific success conditions are weakly specified or absent.
    \item Rewards merely asserting attributes (``quiet,'' ``fast Wi-Fi,'' ``no fees'') without requiring evidence, linkage to the task, or checks against duplication.
    \item Counting proxies dominate while key prompt requirements have only weak gates (e.g., no requirement that citations support specific claims; no requirement that items be distinct and on-topic).
\end{itemize}

\textbf{Do NOT apply when:}
\begin{itemize}
    \item Quantity minimums are paired with robust quality controls that make padding ineffective (e.g., each item must be non-duplicative, tied to a specific claim or user need, and verifiably grounded/bounded).
    \item The main issue is that the rubric is generic and doesn't discriminate at all (use \emph{Low Signal}).
    \item The main issue is a specific criterion that shifts the task or overconstrains acceptable answers (use \emph{Misaligned or Rigid}).
\end{itemize}

\textbf{Illustrative examples:}

\begin{small}
\begin{tabularx}{\linewidth}{@{} l X X @{}}
\toprule
& \textbf{Input Context} & \textbf{Rubric} \\
\midrule
\textsc{Pass} & Provide 5 study tips. & \textit{1 pt each for 5 tips that are (a) non-duplicative and (b) each includes a concrete example of how to apply it.} \\
\addlinespace
\textsc{Fail} & Provide a recommendation. & \textit{5 pts: At least 10 pros. 5 pts: At least 10 cons. (No check for relevance or duplication.)} \\
\bottomrule
\end{tabularx}
\end{small}

\commentout{{\paragraph{Low Signal.}
\label{fm:low_signal}
Apply when the rubric as a whole does not discriminate candidate responses well for this prompt---i.e., it would give similar (often high) scores to many substantively different-quality responses---because the criteria are all generic, conditionally irrelevant, or too easy.

\textbf{How to determine} (rubric-level discrimination test):
\begin{itemize}
    \item Imagine 3--5 candidate responses ranging from weak to excellent.
    \item Ask: Would the rubric's criteria/weights produce nearly equivalent different scores across them based on prompt-relevant success?
    \item If the rubric would likely award similar scores because most criteria are low-signal (e.g., ``helpful,'' ``nice formatting,'' ``completed the task'') and there are few/no strong quality gates tied to the prompt's real success conditions, apply.
\end{itemize}

\textbf{Common signals:}
\begin{itemize}
    \item Most points are allocated to generic writing quality/tone/formatting that is not central for this prompt.
    \item Criteria are trivially satisfied by any minimally on-task response (e.g., ``schedule exists'').
    \item The rubric could be pasted into many unrelated tasks with little or no change.
\end{itemize}

\textbf{Do NOT apply when:}
\begin{itemize}
    \item The rubric is instead missing prompt-imposed must-haves (use \emph{Missing Criteria}).
    \item The rubric imposes the wrong constraints/assumptions (use \emph{Misaligned or Rigid}).
    \item The rubric is gameable specifically via quantity/proxies (use \emph{Hackable}).
\end{itemize}
}

\commentout{\textbf{Illustrative examples:}

\begin{tabularx}{\linewidth}{@{} l X X @{}}
\toprule
& \textbf{Input Context} & \textbf{Rubric} \\
\midrule
\textsc{Pass} & Return JSON only with required keys. & \textit{6 pts: Valid JSON with required keys. 4 pts: No extra text outside JSON.} \\
\addlinespace
\textsc{Fail} & Return only a SQL query. & \textit{5 pts: Response is helpful. 5 pts: Uses appropriate tone.} \\
\bottomrule
\end{tabularx}}


}

\paragraph{Redundant Criteria.}
\label{fm:redundant}
Apply when two or more rubric criteria substantially evaluate the same underlying requirement such that the same behavior is rewarded/penalized multiple times.

\textbf{How to determine:}
\begin{itemize}
    \item Additional-signal test (primary): If these criteria are separate, do you get genuinely different evaluation signal, or are you just re-awarding the same property?
    \item Remove-one test: If removing a criterion would not meaningfully change what is evaluated (only point allocation), it is redundant.
    \item Includes near-duplicates and cases where one criterion fully subsumes another.
\end{itemize}

\textbf{Important clarification} (do NOT apply for mere dependencies):
\begin{itemize}
    \item Do NOT apply just because criteria are related or one tends to enable another.
    \item If Criterion B is a prerequisite/necessary condition for Criterion A (or vice versa) but still measures a distinct dimension (e.g., ``valid JSON'' and ``has required keys''; ``code compiles'' and ``passes tests''), that is NOT redundancy.
\end{itemize}

Do NOT apply when criteria are related but clearly distinct checks (e.g., factual accuracy vs clarity; format compliance vs correctness; presence of citations vs whether citations support claims).

\textbf{Illustrative examples:}

\begin{small}
\begin{tabularx}{\linewidth}{@{} l X X @{}}
\toprule
& \textbf{Input Context} & \textbf{Rubric} \\
\midrule
\textsc{Pass} & Write a research summary with citations. & \textit{3 pts: Claims are supported by citations. 2 pts: Writing is well-organized. 2 pts: Includes limitations of the evidence.} \\
\addlinespace
\textsc{Fail} & Essay rubric. & \textit{5 pts: Clear writing. 5 pts: Clarity of prose. 5 pts: Writing is easy to understand.} \\
\bottomrule
\end{tabularx}
\end{small}

\subsection{Controlled Intervention Operators}
\label{app:rift_interventions}

Each intervention begins from a clean rubric with $M$ criteria.
For a criterion-level arm with requested corruption rate $\rho$, we select
$k=\lfloor M\rho+0.5\rfloor$ criteria without replacement using a seed fixed by
the instance and arm.
A failure-mode arm applies one operator to all selected criteria; a
dimension-level arm assigns selected criteria as evenly as possible across
that dimension's operators.
The clean rubric is retained as the $\rho=0$ condition.

\paragraph{Reliability operators.}
\begin{itemize}
    \item \textbf{Subjective Language} ($1\!\rightarrow\!1$, LLM rewrite):
    remove objective expectations, quantities, and verifiable conditions while
    preserving the criterion's topic, replacing them with unanchored evaluative
    language.
    \item \textbf{Ungrounded Verification} ($1\!\rightarrow\!1$, LLM rewrite):
    preserve the demand for factual verification but remove the reference
    values, sources, tolerances, time scope, or verification procedure needed
    to perform it.
    \item \textbf{Non-Atomic} ($2\!\rightarrow\!1$, LLM merge):
    merge two criteria into one conjunctive pass--fail statement with no
    separable partial-credit rule. Selected criteria are paired; when an odd
    selected criterion remains, an otherwise untouched criterion is sampled as
    its partner when available.
\end{itemize}

\paragraph{Content Validity operators.}
\begin{itemize}
    \item \textbf{Missing Criteria} ($1\!\rightarrow\!0$, deterministic):
    delete each selected criterion. At least one criterion is retained so that
    the corrupted rubric remains scorable.
    \item \textbf{Rigid} ($1\!\rightarrow\!1$, LLM rewrite):
    preserve the underlying requirement but add an exact phrasing, format,
    structure, order, length, or method not required by the task.
    \item \textbf{Redundant Criteria} ($1\!\rightarrow\!2$, LLM duplication):
    retain the selected criterion and insert a logically equivalent paraphrase
    with the same pass--fail boundary.
    \item \textbf{Misaligned} ($1\!\rightarrow\!1$, LLM rewrite):
    replace the selected criterion with a superficially related criterion that
    evaluates an unrequested objective or embeds an incorrect task assumption.
    \item \textbf{Hackability} ($1\!\rightarrow\!1$, LLM rewrite):
    replace the substantive decision rule with an easily inflated surface
    proxy, such as a keyword, heading, response length, or count threshold.
    \item \textbf{Miscalibrated Weighting} ($M\!\rightarrow\!M$,
    deterministic): define core criteria as those with panel
    $\mathrm{CVR}_i>0$, scale their weights down by a common factor, and
    redistribute the removed mass across non-core criteria in proportion to
    their original weights. Here $\rho$ is the requested fraction of total
    weight mass moved; criterion text and total weight remain unchanged.
\end{itemize}

Miscalibrated Weighting is evaluated only as a failure-mode arm because its
corruption rate is measured in weight mass, whereas dimension-level allocation
operates over criteria.
The operator is skipped when the rubric has non-positive weights, an unusable
panel, no core criteria, or no non-core criteria.

\paragraph{Model and validation contract.}
The five rewrite operators, Non-Atomic, and Redundant Criteria receive the task,
the complete indexed rubric, and only the selected indices or index pairs.
They may modify only those targets and must return a fixed JSON object containing
exactly one output per target.
Malformed, missing, duplicate, or off-target outputs are returned to the model
for a bounded repair attempt.
Missing Criteria and Miscalibrated Weighting require no model call.
After transformation, weights are normalized to preserve the clean rubric's
total scoring mass; for a Non-Atomic merge the merged criterion receives the
sum of its source weights, while a redundant duplicate initially receives its
source criterion's weight.

\paragraph{Intervention records.}
For every corrupted rubric, we store the requested arm and $\rho$, seed, clean
and corrupted criterion counts, targeted indices, each before--after
transformation, and the realized corruption rate.
For criterion-level operators, realized corruption is the fraction of original
criteria touched; it can exceed requested $\rho$ when Non-Atomic requires an
additional merge partner.
For Miscalibrated Weighting, realized corruption is the fraction of total weight
mass moved and is capped when necessary to keep all core weights positive.
The model prompt templates are given in \appref{app:corruption_prompts}.

\section{Prompts}
\label{app:prompts}

This section provides the prompt templates used in the RIFT pipeline. 
Template variables are denoted with double curly braces (e.g., \texttt{\{\{variable\}\}}).

\commentout{\subsection{LLM-as-a-Judge Prompt}
\label{app:llmaj_prompt}

TODO}

\subsection{Taxonomy Refinement Prompt}
\label{app:refinement_prompt}

The following prompt template drives the iterative taxonomy refinement process described in Section~\ref{sec:rift} (\tabref{tab:rift_taxonomy}).
During each iteration of the grounded theory pipeline, expert annotator feedback---comprising open-ended rubric critiques and taxonomy critiques---is batched and provided to GPT-5.2 alongside the current taxonomy state.
The model proposes refinements (merges, additions, clarifications, splits, removals, or renames), which are then reviewed and finalized by the expert panel.

\begin{framed}\small
\texttt{You are an expert at analyzing rubric quality feedback and refining failure mode taxonomies. Your task is to output a complete refined failure mode taxonomy.}

\medskip
\texttt{\#\# Original Failure Mode Taxonomy}

\medskip
\texttt{This is the original taxonomy before any refinements in this session:}\\
\textit{[For each failure mode: label and description. If none defined: ``No failure modes have been defined yet.'']}

\medskip
\texttt{\#\# Current Running Refinement}

\medskip
\texttt{This is the taxonomy as refined so far in this session (may be identical to original if this is the first batch):}\\
\textit{[For each failure mode: label, description, rationale, and counts of pass/fail examples. If none refined: ``No refinements have been made yet.'']}

\medskip
\texttt{\#\# Annotator Feedback to Analyze}

\medskip
\texttt{Below are annotations with two types of critiques:}\\
\texttt{- Rubric Critique: Issues the annotator observed in the rubric that were NOT captured by the original taxonomy labels (may suggest new failure modes)}\\
\texttt{- Taxonomy Critique: Critique of the ORIGINAL taxonomy (unclear definitions, overlapping categories, missing categories, etc.). Note: these critiques were written against the original taxonomy, not the running refinement.}

\medskip
\textit{[For each annotation in the batch:]}

\medskip
\texttt{Annotation \{\{loop.index\}\}}\\
\texttt{Input Context: \{\{item.input\_context\}\}}\\
\texttt{Rubric: \{\{item.rubric\}\}}\\
\texttt{Rubric Critique: (issues not captured by original taxonomy)}\\
\texttt{\{\{item.rubric\_critique or "None provided"\}\}}\\
\texttt{Taxonomy Critique: (critique of the original taxonomy)}\\
\texttt{\{\{item.failure\_mode\_critique or "None provided"\}\}}

\medskip
\textit{[End of annotation loop]}
\end{framed}

The prompt further specifies a detailed \textbf{taxonomy philosophy} and \textbf{refinement guidelines} that constrain the model's proposed changes:

\begin{framed}\small
\texttt{\#\# Taxonomy Philosophy}

\medskip
\texttt{CRITICAL: This taxonomy will be used by human annotators. The primary goal is to create a taxonomy that is:}
\begin{itemize}
    \setlength{\itemsep}{1pt}
    \setlength{\parskip}{0pt}
    \item \texttt{Compact}: Aim for 7--10 total failure modes. Fewer distinct categories is ALWAYS better than many granular ones.
    \item \texttt{Easily distinguishable}: A human should be able to distinguish between any two failure modes in under 30 seconds. If two categories require careful reading to tell apart, they should be merged or their distinction should be clarified by refining the label names and or the description.
    \item \texttt{Actionable}: Each category must be clearly applicable without ambiguity.
\end{itemize}

\texttt{Consolidation over proliferation: When in doubt, MERGE rather than add. Two failure modes that are 80\% similar should become one category, not two. The cost of a slightly imperfect merge is far lower than the cost of a bloated, hard-to-use taxonomy.}

\medskip
\texttt{\#\# Guidelines}

\begin{itemize}
    \setlength{\itemsep}{1pt}
    \setlength{\parskip}{0pt}
    \item \texttt{Clear descriptions}: Each failure mode description must be clear, specific, and actionable. The description should explicitly specify HOW to determine if a rubric exhibits this failure mode. An annotator should be able to read the description and confidently apply it to any rubric.
    \item \texttt{No overlapping failure modes}: The taxonomy should not contain failure modes with overlapping meanings. If two labels capture the same concept, merge them or refine them to make them distinct. Do NOT add a new failure mode if its meaning already exists under a different label.
    \item \texttt{Self-contained rationales}: Each rationale must be a self-contained justification that will be used for manual review. It should explain WHY this failure mode exists, what evidence from critiques supports it, and how it differs from other failure modes. A reviewer should understand the rationale without needing to see the original critiques.
    \item \texttt{Cumulative applicability}: The refined taxonomy must be applicable to ALL critiques that have been seen in this session (including previous batches), not just the current batch. Do not remove or change failure modes in ways that would make them inapplicable to earlier critiques that supported them.
\end{itemize}

\medskip
\texttt{\#\# Task}

\medskip
\texttt{Analyze BOTH the rubric critiques and taxonomy critiques above. Before adding any new failure modes, first consider whether existing categories should be merged.}

\medskip
\texttt{FIRST: Consider merging existing failure modes when:}
\begin{itemize}
    \setlength{\itemsep}{1pt}
    \setlength{\parskip}{0pt}
    \item \texttt{Two or more categories have similar descriptions or capture closely related issues}
    \item \texttt{Categories are difficult to distinguish without careful reading}
    \item \texttt{A broader category could capture multiple narrower ones without losing important distinctions}
    \item \texttt{The taxonomy has grown beyond 12 failure modes}
\end{itemize}

\texttt{PREFERRED action - merge: Combine overlapping, redundant, or closely related labels into one. This is the most important refinement action. If you're unsure whether two categories are distinct enough, merge them.}

\medskip
\texttt{Add new failure modes ONLY when ALL of the following are true:}
\begin{itemize}
    \setlength{\itemsep}{1pt}
    \setlength{\parskip}{0pt}
    \item \texttt{The issue is clearly NOT capturable by ANY existing failure mode (even with minor rewording)}
    \item \texttt{The issue appears in MULTIPLE critiques (not just one annotation)}
    \item \texttt{The new category is easily distinguishable from ALL existing categories}
    \item \texttt{Adding it would NOT push the taxonomy beyond 12 failure modes}
\end{itemize}

\texttt{Other refinement actions:}
\begin{itemize}
    \setlength{\itemsep}{1pt}
    \setlength{\parskip}{0pt}
    \item \texttt{clarify: Make a label's description clearer, more specific, or more actionable (especially clarifying HOW to identify the failure mode)}
    \item \texttt{split: Divide an overly broad label into more specific ones (use sparingly---only when a category is genuinely too broad to apply consistently)}
    \item \texttt{remove: Eliminate labels that are not useful, are duplicates, or are too similar to other categories}
    \item \texttt{rename: Change a label name to be more descriptive}
\end{itemize}

\texttt{Output:}\\
\texttt{1. failure\_modes: The complete list of failure modes after applying changes. Each failure mode should have:}
\begin{itemize}
    \setlength{\itemsep}{1pt}
    \setlength{\parskip}{0pt}
    \item \texttt{label: concise identifier (e.g., contradictory\_criteria, missing\_edge\_cases)}
    \item \texttt{description: clear, specific, and actionable description that explains HOW to determine if a rubric has this failure mode (what to look for, what conditions must be met)}
    \item \texttt{rationale: a self-contained justification for this failure mode that can be understood without seeing the original critiques. Explain why it exists, what patterns it captures, and how it differs from related failure modes. If this is a NEW category, explicitly explain why it cannot be captured by any existing category.}
    \item \texttt{examples: REQUIRED: 3-5 pass\_examples AND 3-5 fail\_examples for each failure mode. Multiple diverse examples are essential for annotator training. You may use real examples from the annotations or synthesize clear illustrative examples. Each example should illustrate a distinct scenario or nuance.}
\end{itemize}
\texttt{2. changes\_summary: A list of strings describing what changes you made (e.g., "Added `contradictory\_criteria' based on rubric critiques", "Clarified description of `ambiguous\_criterion'", "Merged `x' and `y' into `z'")}

\medskip
\texttt{If no changes are needed based on these critiques, return the current running refinement unchanged with an empty changes\_summary.}
\end{framed}

\subsection{Controlled Intervention Prompts}
\label{app:corruption_prompts}

The following templates specify the model-backed operators.
The task, full indexed rubric, and target indices are supplied in the user
message:

\begin{framed}\small
\texttt{The task the rubric evaluates:}\\
\texttt{---}\\
\texttt{\{\{instruction\}\}}\\
\texttt{---}

\medskip
\texttt{The rubric (\{\{num\_criteria\}\} criteria):}\\
\texttt{\{\{index\}\}: \{\{criterion\}\}}\\
\textit{[Repeated for every criterion.]}

\medskip
\texttt{Target indices: \{\{target\_indices\}\}}\\
\textit{[For Non-Atomic: \texttt{Target pairs to merge:
\{\{target\_pairs\}\}}.]}
\end{framed}

Each system prompt identifies one RIFT failure mode, includes its taxonomy
definition, and instructs the model to act as a research tool that injects that
known defect so its downstream effects can be measured.
All prompts prohibit changes outside the supplied targets and prohibit
mentioning the rubric, transformation, or instructions in criterion text.
The operator-specific instructions are:

\paragraph{Subjective Language.}
\begin{framed}\small
\texttt{Rewrite each target criterion so that it evaluates roughly the same
aspect of the response but through inherently subjective, unanchored language:
replace objective expectations, named quantities, exact values, and verifiable
conditions with subjective evaluative terms (``clear'', ``appropriate'',
``well-written'', ``high quality'', ``sufficiently detailed'', ``reasonable'').
The rewritten criterion must remain a single true/false statement about the
response, but one whose verdict now depends on the grader's taste. Keep each
rewrite on the same topic as the original---degrade its precision, not its
subject.}
\end{framed}

\paragraph{Ungrounded Verification.}
\begin{framed}\small
\texttt{Rewrite each target criterion by stripping its grounding details while
keeping the demand for factual verification: remove reference values, expected
answers, named sources, formulas, tolerances, time scopes, and verification
procedures, so the criterion still asks the grader to check correctness but no
longer says against what. Keep each rewrite a single true/false statement
demanding factual verification---do not turn it into a subjective quality
judgment.}
\end{framed}

\paragraph{Non-Atomic.}
\begin{framed}\small
\texttt{Merge each target pair into ONE bundled criterion that demands both
original conditions at once, entangled so they cannot be scored separately:
join the requirements with conjunctions or clause chains (``and'', ``while
also'', ``as well as'') into a single pass/fail statement where partial
fulfillment has no consistent verdict. Preserve the substance of both
originals---the defect is bundling, not lost content.}
\end{framed}

\paragraph{Rigid.}
\begin{framed}\small
\texttt{Replace each target criterion with an unjustifiably rigid version of
the same underlying requirement: demand one exact phrasing, format, structure,
ordering, length, or method where the task permits many valid ones. A response
could fully satisfy the task and still fail the criterion on the arbitrary
constraint. Keep the original requirement recognizable and keep each
replacement a single, judgeable true/false statement.}
\end{framed}

\paragraph{Redundant Criteria.}
\begin{framed}\small
\texttt{For each target criterion, write a paraphrased duplicate that evaluates
exactly the same underlying requirement with the same pass/fail boundary, but
is worded differently enough that the duplication is not obvious from a casual
read. The duplicate will be inserted alongside the original, so the same
behavior is scored twice. Any response passing one must pass the other.}
\end{framed}

\paragraph{Misaligned.}
\begin{framed}\small
\texttt{Replace each target criterion with a misaligned version of its
requirement: grade an objective the task did not ask for, or embed a factually
or contextually incorrect assumption about the task. The replacement should
stay superficially on-topic and professional. The misalignment must be
substantive: a grader following it would reward or penalize the wrong thing for
this task. Keep each replacement a single, judgeable true/false statement.}
\end{framed}

\paragraph{Hackability.}
\begin{framed}\small
\texttt{Rewrite each target criterion so its decision rule keys on an easily
inflated surface proxy instead of the underlying requirement: presence of
keywords or section headers, response length or count thresholds, use of
hedging or citation-like formatting, or mentioning a term rather than being
correct about it. A responder who games the proxy should pass without actually
satisfying the original requirement. Keep the proxy superficially related to
the original and express it as a single, judgeable true/false statement.}
\end{framed}

The five rewrite operators and Redundant Criteria return:
\begin{framed}\small
\texttt{\{"rewrites": [\{"index": <target index>, "criterion":
"<rewritten criterion>"\}, ...]\}}
\end{framed}
Non-Atomic returns:
\begin{framed}\small
\texttt{\{"merges": [\{"indices": [<index a>, <index b>], "criterion":
"<merged criterion>"\}, ...]\}}
\end{framed}
The response must contain exactly one entry per target and no prose or code
fences.
Missing Criteria and Miscalibrated Weights are deterministic and therefore
have no model prompt.

\section{Cross-Judge Robustness Check}
\label{app:cross_judge}

\S\ref{sec:res-signatures}'s loudest signatures hackable, rigid, and
misaligned, all driven by a mean CVR collapse and are scored throughout by
a single judge, GPT-5.5. To rule out the effect being an artifact of that
particular scorer, we rescored the clean baseline and each of these three
modes at $\rho{=}0.5$ (240 rubrics total) with an independent 3-judge
panel a \texttt{claude-opus-4-8}, a \texttt{claude-haiku-4-5}, and
a \texttt{gemini-3.1-pro}, sharing no model or provider with GPT-5.5. We
scored 716 of 720 planned units (99.4\%; the remainder failed on a
reasoning-budget limit or a malformed tool call and were not retried
further).

Table~\ref{tab:crossjudge} compares the paired Cohen's $d$ (variant $-$
clean) obtained under this panel against the GPT-5.5 values already
reported in Table~\ref{tab:signatures_full}. All three effects replicate in
direction and in comparable magnitude, each significant at
$p{<}10^{-20}$: the mean CVR collapse under hackable, rigid, and misaligned
is not an artifact of the scoring judge.

\begin{table}[h]
\centering
\small
\caption{Paired Cohen's $d$ for the mean CVR collapse (variant $-$ clean, $n{=}60$/mode) under GPT-5.5 versus an independent 3-judge panel (\texttt{claude-opus-4-8}, \texttt{claude-haiku-4-5}, \texttt{gemini-3.1-pro}).}
\label{tab:crossjudge}
\begin{tabular}{@{}lrrr@{}}
\toprule
\textbf{Mode} & \textbf{GPT-5.5} & \textbf{3-judge panel} & \textbf{$p$} \\
\midrule
Hackable & $-2.83$ & $-2.51$ & $2.6\times10^{-27}$ \\
Rigid & $-2.04$ & $-1.89$ & $2.9\times10^{-21}$ \\
Misaligned & $-1.85$ & $-2.24$ & $8.5\times10^{-25}$ \\
\bottomrule
\end{tabular}
\end{table}


\section{Full Per-Mode Signatures}
\label{app:signatures}

Table~\ref{tab:signatures_full} reports every FDR-significant signal for each
of the eight criterion-level failure modes at $\rho{=}0.5$, expanding the
heatmap of Figure~\ref{fig:heatmap} into exact effect sizes. Modes are grouped
by the family the data produces rather than by a grouping we imposed: the
misspecification modes collapse content validity while leaving criterion
counts intact, the structural modes move counts while content validity holds,
and the ambiguity modes produce only a faint relevance dip.

\begin{table}[h]
\centering
\small
\setlength{\tabcolsep}{5pt}
\renewcommand{\arraystretch}{1.1}
\caption{Per-mode signatures at $\rho{=}0.5$ ($n{=}60$ per mode). Paired
Cohen's $d$ (variant $-$ baseline); all entries shown are FDR-significant at
$q<.05$ over the full $8\times17$ grid. $k$ is the number of significant
signals for that mode. AUC is the unpaired Mann--Whitney
$P(\text{variant} > \text{baseline})$, where $0.5$ indicates a signal blind to
the corruption.}
\label{tab:signatures_full}
\begin{tabularx}{\textwidth}{@{}llrrX@{}}
\toprule
\textbf{Mode} & \textbf{$k$} & \textbf{Signal} & \textbf{$d$} & \textbf{Note} \\
\midrule

\multicolumn{5}{@{}l}{\emph{Misspecification: content validity collapses, criterion counts unchanged}} \\
\addlinespace[2pt]

\multirow{10}{*}{Hackable} & \multirow{10}{*}{10}
  & mean\_cvr              & $-2.83$ & AUC $.13$, $p{=}5\times10^{-30}$; loudest effect in the corpus \\
& & weighted\_relevance    & $-2.07$ & \\
& & item\_relevance        & $-1.94$ & \\
& & candidate\_item\_mean  & $-1.94$ & \\
& & core\_pct              & $-1.17$ & AUC $.29$ \\
& & general\_pct           & $+0.72$ & Mass drifts outward from the task domain \\
& & irrelevant\_pct        & $+0.46$ & \\
& & num\_flags             & $+0.41$ & \\
& & background\_pct        & $+0.38$ & \\
& & representativeness     & $-0.38$ & \\
\cmidrule(l){1-5}

\multirow{7}{*}{Rigid} & \multirow{7}{*}{7}
  & mean\_cvr              & $-2.04$ & AUC $.15$ \\
& & weighted\_relevance    & $-0.55$ & \\
& & core\_pct              & $-0.39$ & \\
& & item\_relevance        & $-0.38$ & \\
& & candidate\_item\_mean  & $-0.38$ & \\
& & representativeness     & $-0.38$ & \\
& & general\_pct           & $+0.35$ & \\
\cmidrule(l){1-5}

\multirow{6}{*}{Misaligned} & \multirow{6}{*}{6}
  & mean\_cvr              & $-1.85$ & AUC $.18$ \\
& & weighted\_relevance    & $-0.73$ & \\
& & item\_relevance        & $-0.59$ & \\
& & candidate\_item\_mean  & $-0.59$ & \\
& & irrelevant\_pct        & $+0.46$ & Distinguishes misaligned from rigid \\
& & core\_pct              & $-0.43$ & \\
\midrule

\multicolumn{5}{@{}l}{\emph{Structural: criterion counts move, content validity intact}} \\
\addlinespace[2pt]

\multirow{6}{*}{Redundant} & \multirow{6}{*}{6}
  & total\_items           & $+2.30$ & Mean $+7.3$ criteria \\
& & items\_bad             & $+1.41$ & \\
& & items\_review          & $+1.07$ & \\
& & items\_good            & $+0.65$ & \\
& & redundancy\_score      & $+0.64$ & Fires for this mode and no other \\
& & weighted\_relevance    & $+0.31$ & \\
\cmidrule(l){1-5}

\multirow{5}{*}{Non-atomic} & \multirow{5}{*}{5}
  & total\_items           & $-2.47$ & Mean $-3.9$ criteria \\
& & weighted\_relevance    & $+1.50$ & Inversion: merged criteria appear \emph{more} relevant \\
& & items\_review          & $-0.90$ & \\
& & items\_good            & $-0.73$ & \\
& & core\_pct              & $+0.32$ & \\
\cmidrule(l){1-5}

\multirow{5}{*}{\makecell[l]{Missing\\criteria}} & \multirow{5}{*}{5}
  & total\_items           & $-2.30$ & Mean $-7.3$ criteria \\
& & items\_bad             & $-1.43$ & \\
& & items\_review          & $-1.22$ & \\
& & items\_good            & $-0.67$ & \\
& & representativeness     & $-0.38$ & Surviving criteria stay individually healthy; only composition shifts \\
\midrule

\multicolumn{5}{@{}l}{\emph{Ambiguity: neither pattern, only a faint relevance dip}} \\
\addlinespace[2pt]

\multirow{5}{*}{\makecell[l]{Subjective\\language}} & \multirow{5}{*}{5}
  & weighted\_relevance    & $-0.67$ & \\
& & item\_relevance        & $-0.62$ & \\
& & candidate\_item\_mean  & $-0.62$ & \\
& & general\_pct           & $+0.46$ & \\
& & core\_pct              & $-0.41$ & mean\_cvr is \emph{not} significant here ($-0.09$) \\
\cmidrule(l){1-5}

\multirow{2}{*}{\makecell[l]{Ungrounded\\verification}} & \multirow{2}{*}{2}
  & weighted\_relevance    & $-0.32$ & $\Delta{=}-0.006$ \\
& & num\_flags             & $-0.31$ & $\Delta{=}-0.22$; all other signals silent \\
\bottomrule
\end{tabularx}
\end{table}

Two signals, \texttt{contamination\_score} and \texttt{adjacent\_pct}, do not
reach significance for any mode and are reported as honest nulls. Ungrounded
verification is the sole mode without a strong signature: its largest effect
is $|d|{=}0.32$, and the classifier of \S\ref{sec:res-signatures} recovers it
at $0.78$ recall only from the joint pattern of a mild relevance dip occurring
\emph{without} the accompanying content-validity collapse.

\clearpage

\section{Corruption-Level Response}
\label{app:dose}

\begin{figure}[h]
\centering
\includegraphics[width=\textwidth]{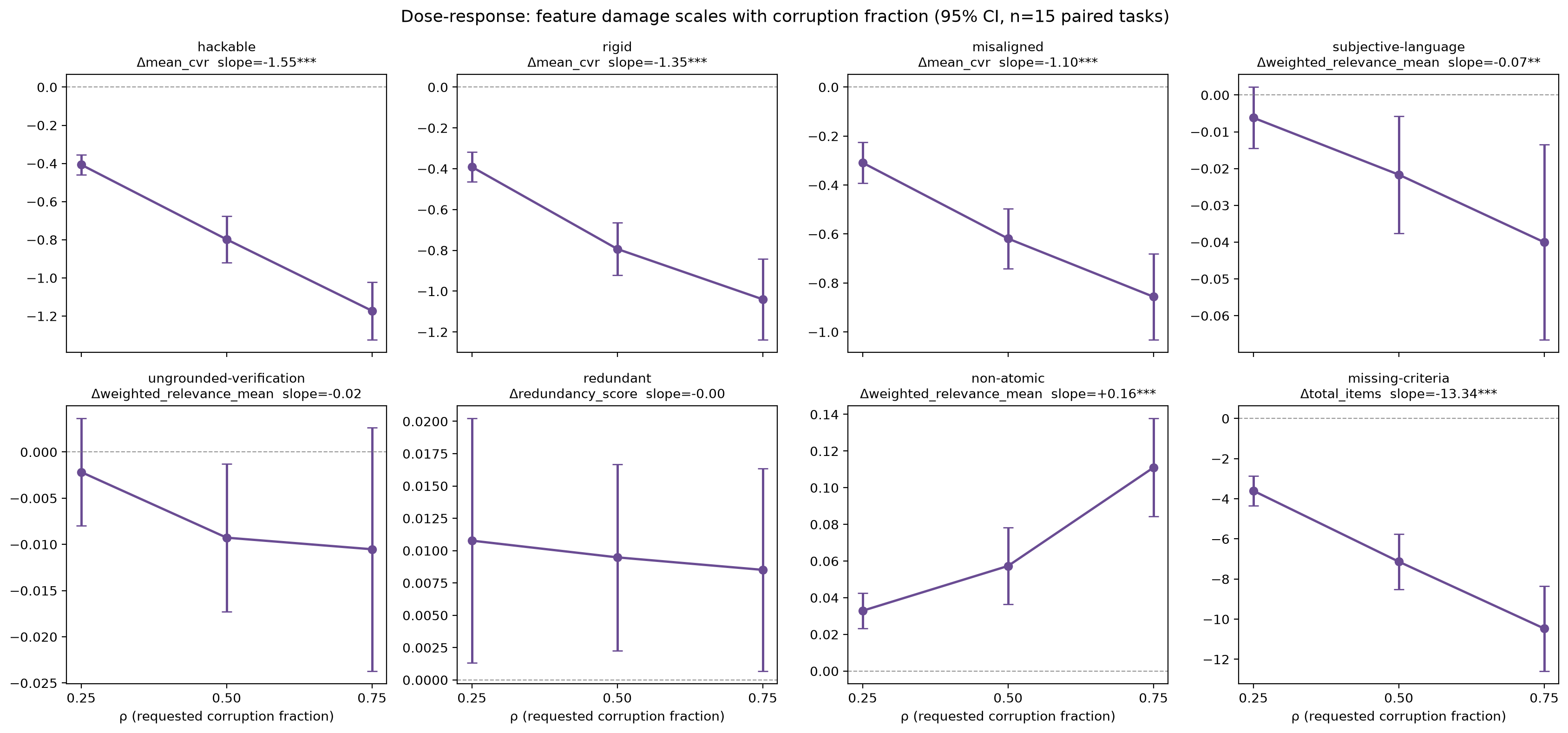}
\caption{Signal response against realized corruption fraction $\rho$
($95\%$ CI, $n{=}15$ paired tasks per level). The CVR-based signals and
\texttt{total\_items} descend steadily with corruption, while
\texttt{redundancy\_score} and ungrounded verification's weighted relevance
stay flat across the full range.}
\label{fig:dose}
\end{figure}

Figure~\ref{fig:dose} and Table~\ref{tab:dose_full} report how each signature
scales with the amount of corruption applied. Every unit is regressed on its
\emph{realized} corruption fraction rather than the requested level, since the
operators do not always achieve the requested fraction exactly.

\begin{table}[h]
\centering
\small
\setlength{\tabcolsep}{6pt}
\renewcommand{\arraystretch}{1.1}
\caption{OLS slope of $\Delta$signal on realized $\rho$ over the 15-task
subset (3 corruption levels, 45 paired points per mode). Per-family slopes
give the same regression fit within the GDPval and Terminal-Bench subsets
independently. Realized corruption fractions average $0.267$, $0.528$, and
$0.766$ at the three requested levels, with non-overlapping ranges.}
\label{tab:dose_full}
\begin{tabular}{@{}llrrrl@{}}
\toprule
\textbf{Mode} & \textbf{Signal} & \textbf{Slope} & \textbf{$r$} & \textbf{$p$} & \textbf{GDPval / TB} \\
\midrule
\multicolumn{6}{@{}l}{\emph{Graded: slope estimates severity}} \\
\addlinespace[2pt]
Hackable            & $\Delta$mean\_cvr           & $-1.55$  & $-.83$ & $1.6\times10^{-12}$ & $-1.83$ / $-1.21$ \\
Rigid               & $\Delta$mean\_cvr           & $-1.35$  & $-.73$ & $1.6\times10^{-8}$  & $-1.62$ / $-1.01$ \\
Misaligned          & $\Delta$mean\_cvr           & $-1.10$  &        & $5\times10^{-7}$    & \\
Missing criteria    & $\Delta$total\_items        & $-13.3$  &        & $<.001$             & Mechanical \\
Non-atomic          & $\Delta$weighted\_relevance & $+0.159$ & $+.67$ & $<.001$             & Inverted \\
Subjective language & $\Delta$weighted\_relevance & $-0.073$ &        & $<.01$              & \\
\addlinespace[4pt]
\multicolumn{6}{@{}l}{\emph{Weakly graded: detects presence, estimates severity poorly}} \\
\addlinespace[2pt]
Hackable            & $\Delta$core\_pct           & $-0.23$  & $-.35$ & $<.05$              & \\
\addlinespace[4pt]
\multicolumn{6}{@{}l}{\emph{Flat: presence only}} \\
\addlinespace[2pt]
Redundant           & $\Delta$redundancy\_score   & $-0.005$ &        & n.s.                & Saturates \\
Ungrounded verif.   & $\Delta$weighted\_relevance & $-0.018$ &        & n.s.                & Stealthy at every level \\
\bottomrule
\end{tabular}
\end{table}

The division is functional rather than following the boundaries between
signal families. Content validity behaves as a \emph{dosimeter} for the
misspecification modes, so the magnitude of the drop estimates how much of the
rubric has been compromised. The domain-mapping signals behave as
\emph{presence detectors}: \texttt{core\_pct} registers that hackability has
been introduced but grades its extent only weakly. The cross-lens signals
catch the modes that invert the logic of the other two, most visibly
non-atomicity, whose weighted relevance climbs as more criteria are merged.
That inversion has a consequence beyond this corpus: any rubric-quality
measure built on mean per-criterion relevance can be improved by consolidating
criteria rather than by improving them.

\clearpage

\commentout{
\section{Weight-Calibration Signals}
\label{app:weightcal}

Table~\ref{tab:weightcal} reports the exact values behind
Figure~\ref{fig:weightcal} and \S\ref{sec:res-weight}: the three
weight-calibration statistics on the 60 human-authored rubrics, and their
response under weight-shuffle injection at each corruption level.

\begin{table}[h]
\centering
\small
\caption{Weight-calibration signals on human rubrics ($n{=}60$, panel-scored) and under weight-shuffle injection ($n{=}35$ paired). Higher $\rho_{\mathrm{ord}}$ and core weight share are better; higher FSR is worse.}
\label{tab:weightcal}
\begin{tabular}{@{}lrrrrr@{}}
\toprule
& \textbf{Human} & \textbf{Clean} & $\rho{=}.25$ & $\rho{=}.50$ & $\rho{=}.75$ \\
\midrule
$\rho_{\mathrm{ord}}$ & $+0.195$ & $+0.189$ & $-0.522$ & $-0.562$ & $-0.562$ \\
Core weight share & $0.835$ & $0.717$ & $0.471$ & $0.246$ & $0.083$ \\
FSR & $0.950$ & $0.956$ & $0.974$ & $0.988$ & $0.996$ \\
\bottomrule
\end{tabular}
\end{table}}

\section{Task-adapted persona panels: examples}
\label{app:personas}

Each task's panel is generated once from the task instruction, never from the
rubric, yielding five one-line expert profiles that are persisted and reused
unchanged across the baseline and every injected condition, and across both the
CVR panel and the cross-lens expert panel (the latter substitutes a neutral base
prompt for the CVR-specific one). Every persona shares the task's base prompt and
differs only in its profile line, so the five raters vary in \emph{vantage point}
while receiving an identical task and response format.

Table~\ref{tab:persona-examples} reproduces the generated panels for two GDPval
and two Terminal-Bench tasks. Two regularities recur across the corpus. First,
panels span complementary roles rather than five near-synonyms: practitioner,
specialist, compliance, audit, and stakeholder perspectives appear consistently,
adapted to the domain. Second, the generator tracks task genre. GDPval
deliverables elicit professional-stakeholder personas, whereas Terminal-Bench
tasks elicit verification-oriented roles keyed to the task's contract, each
phrased around what a rubric must check.

\newcommand{\taskrow}[2]{%
  \multicolumn{2}{@{}p{\linewidth}@{}}{%
    \rule{0pt}{2.6ex}\textbf{#1}\ \ \emph{#2}}\\[2pt]}

\begin{table}[h]
\centering
\footnotesize
\setlength{\tabcolsep}{4pt}
\renewcommand{\arraystretch}{1.15}
\caption{Generated persona panels for four tasks. Each panel is produced once
per task from the instruction alone and reused across all experimental
conditions. Profile lines are verbatim; task descriptions are our one-line
summaries.}
\label{tab:persona-examples}
\begin{tabular}{@{}>{\raggedright\arraybackslash}p{0.24\linewidth}
                  >{\raggedright\arraybackslash}p{0.72\linewidth}@{}}
\toprule
\taskrow{GDPval: EMS cardiac quality-assurance review.}{A paramedic
  reconstructs county EMS STEMI cases for a medical-direction QA report.}
\midrule
EMS medical director & Focused on whether STEMI recognition, alert activation, and protocol adherence are clinically justified by the available field data. \\
Cardiac QA paramedic & A senior QA reviewer assessing reconstructed EMS timelines, ECG timing, documentation consistency, and field decision-making. \\
Interventional cardiologist & Evaluates whether prehospital STEMI suspicion aligns with ECG findings, ED assessment, and cath lab outcomes. \\
Protocol compliance officer & Determines whether ECG acquisition timing and STEMI alert decisions meet county EMS standards. \\
Documentation auditor & Checks that conclusions are supported by reconciled evidence from PCRs, monitor exports, hospital notes, and protocols. \\
\midrule

\taskrow{GDPval: B2B SaaS release retrospective.}{A QA lead analyses defect
  leakage after Release~4.7 and recommends test-plan changes.}
\midrule
QA retrospective lead & Experienced in post-release defect leakage retrospectives for B2B SaaS products. \\
Test coverage analyst & Focused on mapping requirements, executed test cases, automation inventory, and production defects. \\
Customer impact reviewer & Evaluates whether defect severity and customer impact are accurately represented. \\
Automation strategy manager & Assesses whether proposed automation candidates are justified by escaped defects and implementation effort. \\
VP of quality governance & Judges whether the retrospective is direct, actionable, and credible enough for executive review. \\
\midrule

\taskrow{Terminal-Bench: \texttt{rollsnap} Go CLI.}{Implement a
  standard-library-only CLI producing a deterministic, dependency-first
  deployment snapshot.}
\midrule
CLI contract reviewer & Whether criteria are necessary to verify the CLI follows the specified command-line behavior and documented contract. \\
Deterministic rollout validator & Whether criteria are essential for confirming dependency-first deterministic snapshot generation. \\
Input resilience auditor & Whether criteria are necessary to judge tolerance of recoverable corrupt rollout mirror inputs and issue reporting. \\
Standard-library enforcer & Whether criteria are essential for enforcing standard-library-only implementation and prohibited imports. \\
Output artifact checker & Whether criteria are essential for validating parent-directory creation and exact JSON/TSV output artifacts. \\
\midrule

\taskrow{Terminal-Bench: signed-receipt parser differential (CTF).}{Probe an
  HTTP service under a request budget and forge a signed receipt that verifies
  as authorized.}
\midrule
HTTP probe recon analyst & Whether criteria are necessary for correctly probing the service, using \texttt{/verify} as ground truth, and respecting request-count constraints. \\
Parser disagreement specialist & Whether criteria are necessary for identifying the canonical parser disagreement channel and exploiting divergent claim interpretation. \\
Receipt forgery reviewer & Whether criteria are necessary for producing a forged signed receipt that verifies as authorized while parsers disagree. \\
Output schema validator & Whether criteria are necessary for writing stage JSON files with exact key order, digest formats, and required paths. \\
Grader compliance auditor & Whether criteria are necessary for satisfying grader-visible constraints such as action-log limits, required probes, and persistence across stages. \\
\bottomrule
\end{tabular}
\end{table}

\end{document}